  \RequirePackage{fix-cm}
  \documentclass[natbib,twocolumn]{svjour3}          
  \smartqed  
\usepackage{pifont}
\usepackage{amsmath,amsfonts}
\usepackage{algorithmic}
\usepackage{algorithm}
\usepackage{array}
\usepackage{microtype}
\usepackage{xspace}
\usepackage[caption=false,font=normalsize,labelfont=sf,textfont=sf]{subfig}
\usepackage{textcomp}
\usepackage{stfloats}
\usepackage{url}
\usepackage{verbatim}
\usepackage{graphicx}
\usepackage{multirow}
\newcolumntype{Y}{>{\centering\arraybackslash}X}
\newcolumntype{H}{>{\setbox0=\hbox\bgroup}c<{\egroup}@{}}
\usepackage{graphicx}
\usepackage{amsmath}
\usepackage{amssymb}
\usepackage{booktabs}
\usepackage{mathtools}
\usepackage[table]{xcolor}
\usepackage[pagebackref=true,breaklinks=true,citecolor=tealblue,colorlinks,linkcolor=ruby,bookmarks=false]{hyperref}
\definecolor{ruby}{rgb}{0.88, 0.07, 0.37}
\definecolor{tealblue}{rgb}{0.18, 0.40, 0.46}

\begin{document}

\title{One-shot Generative Domain Adaptation in 3D GANs}

\author{Ziqiang Li$^{1,2,*}$ \and Yi Wu$^{2,*}$ \and Chaoyue Wang$^{3}$ \and Xue Rui$^{1}$ \and Bin  Li$^{2,4}$}

\institute{
   {\textit{Corresponding Author:} Chaoyue Wang, Bin Li} \at \email{chaoyue.wang@outlook.com, binli@ustc.edu.cn} \and
  $^*$ These authors contributed equally to this work.\\
  $^1$ Nanjing University of Information Science and Technology, Nanjing, China. \\
  $^2$ University of Science and Technology of China, Hefei, China. \\
  $^3$ University of Sydney, Australia.\\
  $^4$ CAS Key Laboratory of Technology in Geo-spatial Information Processing and Application System, University of Science and Technology of China, Hefei, China.
}

\date{Received: October 7, 2023 / Accepted: xxx }

\def\smallgap{\vspace{0.05in}}
\maketitle
\begin{abstract}
3D-aware image generation necessitates extensive training data to ensure stable training and mitigate the risk of overfitting. This paper first considers a novel task known as One-shot 3D Generative Domain Adaptation (GDA), aimed at transferring a pre-trained 3D generator from one domain to a new one, relying solely on a single reference image. One-shot 3D GDA is characterized by the pursuit of specific attributes, namely, \textit{high fidelity}, \textit{large diversity}, \textit{cross-domain consistency}, and \textit{multi-view consistency}. Within this paper, we introduce 3D-Adapter, the first one-shot 3D GDA method, for diverse and faithful generation. Our approach begins by judiciously selecting a restricted weight set for fine-tuning, and subsequently leverages four advanced loss functions to facilitate adaptation. An efficient progressive fine-tuning strategy is also implemented to enhance the adaptation process. The synergy of these three technological components empowers 3D-Adapter to achieve remarkable performance, substantiated both quantitatively and qualitatively, across all desired  properties of 3D GDA. Furthermore, 3D-Adapter seamlessly extends its capabilities to zero-shot scenarios, and preserves the potential for crucial tasks such as interpolation, reconstruction, and editing within the latent space of the pre-trained generator. Code will be available at \href{https://github.com/iceli1007/3D-Adapter}{https://github.com/iceli1007/3D-Adapter}.
\end{abstract}

\section{Introduction}\label{sec1}

The realm of image generation has witnessed significant advancements, owing to the evolution of deep generative models. These models encompass Variational Autoencoders (VAEs) \cite{girin2020dynamical,zhao2017infovae}, Generative Adversarial Networks (GANs) \cite{goodfellow2020generative,karras2019style,karras2020analyzing,li2022new,li2023systematic,li2023exploring,tao2019resattr}, and Diffusion models \cite{ho2020denoising, nichol2021improved,ruiz2023dreambooth,wu2024infinite}. Notably, there has been a recent surge in efforts to extend these 2D image generation capabilities to the domain of 3D-aware image generation. This expansion involves integrating rendering techniques with neural scene representation, enabling the synthesis of 2D images while concurrently learning 3D structures without explicit 3D supervision. This innovative training paradigm allows 3D generators to produce highly realistic images with consistent multi-view representations, thereby significantly enhancing the scope and potential of generative models.

Similar to 2D generative models \cite{li2022fakeclr，li2022comprehensive}, 3D generative models \cite{chan2022efficient,or2022stylesdf,zhao2022generative,skorokhodov2022epigraf,gu2021stylenerf} require large-scale training data to ensure training stability and mitigate the risk of overfitting. When training data is limited, these models often suffer significant performance degradation, affecting both texture realism and 3D geometry consistency. Unfortunately, there are scenarios where acquiring sufficient training data is impractical. This paper addresses the challenge of one-shot 3D Generative Domain Adaptation (GDA), a task that involves transferring a pre-trained 3D generator from one domain to a new domain using only a single reference image. In this endeavor, we leverage the inherited generative capabilities of the pre-trained model to achieve 3D-aware image generation with four key properties:  \textit{(i)} \textit{High fidelity.} The synthesized images should seamlessly integrate into the target domain, aligning with the attributes of the one-shot reference image.
\textit{(ii)} \textit{Large diversity.} The adapted generator should not merely replicate the one-shot reference image but should exhibit a rich array of variations.
\textit{(iii)} \textit{Cross-domain consistency.} The adapted images and their corresponding source images should maintain consistency in domain-independent attributes, such as pose and identity.
\textit{(iv)} \textit{Multi-view Consistency.} The adapted generator should consistently represent the 3D shape across various views of the same image.

\begin{figure}[t!]
	\centering
	\includegraphics[scale=0.55]{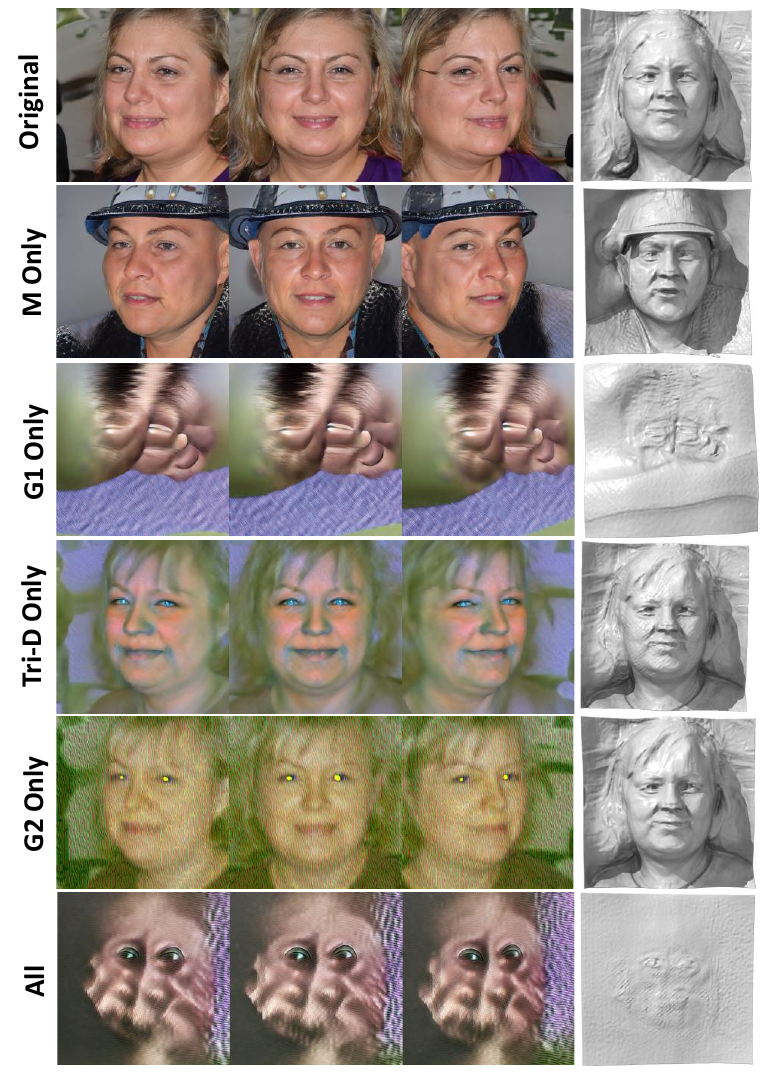}
	\caption{\textbf{Training parameters determination}. We do the ablation study on fine-tuning different components of the EG3D. }
	\label{fig:ablation_module}
\end{figure}

While several studies \cite{abdal20233davatargan,zhang2023styleavatar3d,song2022diffusion} have explored domain adaptation within 3D GANs by leveraging diverse training data in the target domain, the specific area of one-shot GDA remains uncharted within the 3D GAN domain. To address this gap, we present \textbf{3D-Adapter}, which tackles the challenge of using just a single target image to advance 3D-aware image generation in target domains. Our approach strives to achieve all four essential properties: \textit{High fidelity}, \textit{Large diversity}, \textit{Cross-domain consistency}, and \textit{Multi-view consistency}.

Our 3D-Adapter framework builds upon the foundation of the popular 3D GANs, EG3D \cite{chan2022efficient}. EG3D, pre-trained on a large-scale source dataset, excels in delivering reality and view-consistent 3D-aware image synthesis within the source domain. To actualize our method, we have designed three key components: \textit{(i) Determining a Restricted Weight Set for Tuning:} We conducted a comprehensive investigation, exploring models fine-tuned with the original training approach on the source dataset. Our findings revealed that fine-tuning the entire model led to significant performance degradation, affecting both texture and geometry, as exemplified in Figure \ref{fig:ablation_module}. Conversely, selective fine-tuning of specific weight sets, such as Tri-Decoder (Tri-D) and Style-based Super-resolution module (G2), primarily affected texture information while making minimal changes to 3D geometry. Consequently, the fine-tuning of these selected weight sets, Tri-D and G2, emerged as a viable strategy, enhancing training stability and mitigating the challenges associated with one-shot 3D GDA. \textit{(ii) Employing Four Advanced Loss Functions for Adaptation:} Training instability, particularly with adversarial loss, posed a notable challenge. As depicted in Figure \ref{fig:ablation_module}, using adversarial loss alone for fine-tuning either Tri-D or G2 failed to capture the target domain's texture information. To address this issue and achieve the four essential properties of GDA, we introduced four advanced loss functions: domain direction regularization, target distribution learning, image-level source structure maintenance, and feature-level source structure maintenance. Domain direction regularization and target distribution learning leverage the pre-trained Contrastive Language-Image Pre-Training (CLIP) model to ensure substantial diversity and high fidelity, respectively. Additionally, image-level and feature-level source structure maintenance naturally foster cross-domain consistency and facilitate multi-view consistency. \textit{(iii) Implementing an Efficient Progressive Fine-Tuning Strategy:} Directly fine-tuning the Tri-D and G2 modules presented challenges in balancing high fidelity and cross-domain consistency. Therefore, we introduced a two-step fine-tuning strategy that progressively refines the source generator into the target generator. This progressive fine-tuning approach simplifies the process of achieving significant improvements in both high fidelity and cross-domain consistency, further enhancing the effectiveness of our method.

The key contributions of our work can be summarized as follows:

\begin{itemize}
\setlength{\itemsep}{2pt}
\setlength{\parsep}{2pt}
\setlength{\parskip}{2pt}
    \item To the best of our knowledge, this paper is the first exploration of one-shot GDA within the 3D GANs. 
    \item We investigate the impact of fine-tuning various components of the original 3D generator, complemented by the introduction of a progressive fine-tuning strategy. This novel strategy significantly mitigates the challenges associated with one-shot GDA.
    \item we introduce four loss functions, each designed to address the essential aspects of GDA. These loss functions enable the target generator to inherit geometric knowledge from the source generator while accurately capturing the unique texture features of the target domain, utilizing only one reference image. Comprehensive qualitative and quantitative evaluations underscore the impressive performance of our 3D-Adapter across a diverse array of target domains.
\end{itemize}
\section{Related Work}\label{sec2}

\subsection{3D-aware Image Generation.}

  GANs have acquired substantial recognition for their capacity to facilitate 2D image synthesis \cite{karras2019style, karras2020analyzing, goodfellow2014generative, karras2017progressive}. Recent studies \cite{chan2022efficient, or2022stylesdf, zhao2022generative, skorokhodov2022epigraf, gu2021stylenerf} have extended these capabilities to 3D generation by introducing neural scene representation and rendering into generative models. 3D-aware image generation aims to achieve multi-view-consistent image synthesis and extraction of 3D shapes without requiring supervision on geometry or multi-view image collections. Early methods based on voxel-based (explicit) representations \cite{nguyen2019hologan, nguyen2020blockgan} suffered challenges in generating high-resolution content due to the enormous memory demands inherent to voxel grids.

In response to these challenges, recent studies have introduced Neural Radiance Field (NeRF)-based (implicit) representations \cite{mildenhall2021nerf} into the realm of 3D-aware image generation. While these approaches have achieved impressive performance, they are characterized by long query times, leading to inefficiencies in the training process and limiting the attainable degree of realism. To address these inefficiencies, recent studies have proposed hybrid representations \cite{chan2022efficient, devries2021unconstrained} that combine the benefits of both explicit and implicit representations, resulting in architectures that are more efficient in terms of computation and memory. Notable examples of such hybrid architectures include EG3D \cite{chan2022efficient}, which integrates a tri-plane hybrid 3D representation with a StyleGAN2-based framework, and Rodin \cite{wang2023rodin}, which utilizes a diffusion model to create a tri-plane hybrid 3D representation. Our method builds on EG3D \cite{chan2022efficient}, the most popular 3D-aware image generation technology.


Concurrently, emerging researchers have embarked on leveraging pre-trained generative models to advance 3D content generation. A notable example is the Dreamfusion model \cite{poole2022dreamfusion}, which employs Score Distillation Sampling techniques to distill knowledge from a pre-trained 2D text-to-image diffusion model. The primary objective of this approach is to optimize the NeRF for text-to-3D synthesis. However, Dreamfusion relies on a low-resolution diffusion model and a large global MLP for volume rendering, making the approach computationally expensive and prone to performance degradation as image resolution increases. To address these limitations, Magic3D \cite{lin2023magic3d} adopts a two-stage coarse-to-fine framework and utilizes a sparse 3D hash grid structure to enable high-resolution text-to-3D synthesis. This innovation enhances both the efficiency and quality of the generated 3D content, overcoming the drawbacks associated with Dreamfusion.

\begin{figure*}[t!]

	\centering
	\includegraphics[scale=0.65]{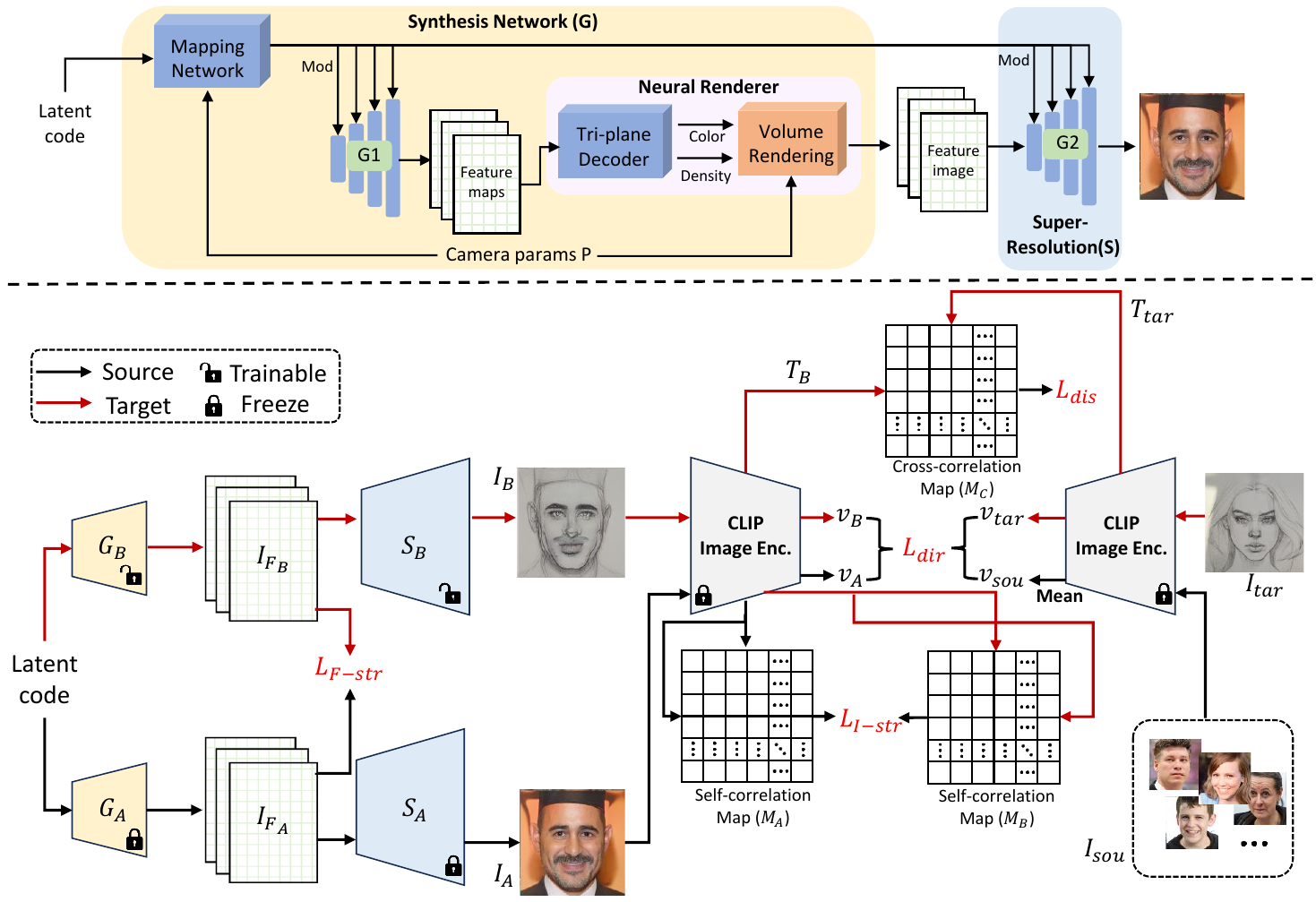}
	\caption{\textbf{The overall generator architecture of EG3D (Top) and our 3D-Adapter (Bottom).} The EG3D consists of two parts: the Synthesis Network, represented by a yellow box, and the Super-Resolution module, represented by a blue box. These correspond to the yellow component $G$ and the blue component $S$ in our 3D-Adapter. The proposed 3D-Adapter is designed to transfer the knowledge from EG3D's generator ($G_A$ and $S_A$), pre-trained on the source dataset, to the target domain ($G_B$ and $S_B$).}
	\label{fig:eg3d}
\end{figure*}

\subsection{Few-shot GDA in 2D GANs.}

Few-shot GDA \cite{ojha2021few,zhao2022closer,zhanggeneralized,zhang2022towards,wu2023domain,alanov2022hyperdomainnet,yang2023one} is focused on the challenge of transferring a pre-trained source generator to a target domain using few reference images, sometimes as scarce as one. Few-shot GDA is underscored by three attributes: \textit{(i)} \textit{High fidelity.} The adapted images should be in the same domain as the few-shot target images. \textit{(ii)} \textit{Large diversity.} The adapted generator should not simply replicate the training images. \textit{(iii)} \textit{Cross-domain consistency.} The adapted images and their corresponding source images should be consistent in terms of domain-sharing attributes.

Nevertheless, the deployment of few-shot GDA is beset by significant challenges arising from limited training data. These challenges manifest as severe overfitting and unstable training processes, which detrimentally impact the diversity and realism of the generative output. To address these issues, recent studies \cite{ojha2021few, zhao2022closer, zhanggeneralized, zhang2022towards} have explored fine-tuning the entire generator, complemented by various regularization techniques. For instance, some studies have introduced a consistency loss based on Kullback-Leibler (KL) divergence \cite{ojha2021few, xiao2022few}, while others have advocated the adoption of contrastive learning methodologies \cite{zhao2022closer} to preserve the relative similarities between the source and target domains. These approaches aim to mitigate the issues of overfitting and instability, thereby enhancing the performance of few-shot GDA models.


Furthermore, leveraging the semantic power of CLIP models \cite{radford2021learning}, some studies \cite{gal2021stylegan, zhang2022towards, zhu2021mind,li2023peer} have proposed a novel approach: defining the domain-gap direction within the CLIP embedding space. This guiding directive steers the optimization process of the target generator towards a domain-consistent outcome. Additionally, some investigations have utilized GAN inversion techniques to reveal domain-sharing attributes between source and target domains \cite{zhang2022towards, zhu2021mind}. Alternatively, they compress the latent space of the target domain into a more compact subspace, offering a solution to the complexities of cross-domain alignment \cite{xiao2022few}. It is noteworthy, however, that these strategies introduce numerous training hyper-parameters, which can sometimes lead to training instability.

Recent studies \cite{alanov2022hyperdomainnet, kim2022dynagan, wu2023domain} have introduced an alternative perspective, suggesting that the efficacy of few-shot GDA can be achieved by strategically freezing the pre-trained source generator. This approach redirects the focus towards training supplementary lightweight re-modulation modules, a paradigm that has demonstrated remarkable proficiency in achieving superior performance in few-shot GDA. This streamlined method not only offers operational efficiency but also ease of implementation.

It is worth noting that, in contrast to 2D GANs, 3D GANs experience more pronounced performance degradation when faced with limited training data \cite{yang2022improving}. To counter this, we advocate for the selective freezing of the majority of parameters within the pre-trained source generator. Fine-tuning efforts are then concentrated on the lightweight Tri-plane decoder and Super-resolution module. This novel strategy has been validated through a series of comprehensive empirical experiments, underscoring its efficacy and applicability in the challenging context of one-shot 3D GDA.

\subsection{Domain Adaptation of 3D GANs.} 

EG3D \cite{chan2022efficient} is a popular 3D-aware image generation method. However, it relies on accurate pose estimation for real-face datasets like FFHQ. Consequently, this method cannot be directly applied to stylized, artistic, or highly variable geometry datasets, where camera and pose information is challenging to estimate. Therefore, many researchers have explored domain adaptation for these datasets. For instance, 3DAvatarGAN \cite{abdal20233davatargan} employs a 2D generator pre-trained on target datasets and a source 3D generator to transfer knowledge from the 2D generator to the 3D generator. Leveraging the score distillation sampling technique proposed in DreamFusion \cite{poole2022dreamfusion}, \cite{song2022diffusion} utilize pre-trained text-to-image diffusion models to adapt a pre-trained 3D generator \cite{chan2022efficient} to a new text-defined domain. StyleAvatar3D \cite{zhang2023styleavatar3d} focuses on calibrating data and efficiently using it to train 3D GANs. Furthermore, some studies \cite{kim2023podia,kim2023datid,song2022diffusion} explore the zero-shot GDA based on EG3D. These methods utilize CLIP \cite{radford2021learning} or text-to-image diffusion models \cite{rombach2022high} pre-trained on a large number of image-text pairs, allowing for text-driven domain adaptation. In comparison to these works, our approach uniquely focuses on using as few as a single reference image to convincingly adapt the source 3D generator to the target domain while maintaining target-domain consistency, large diversity, and cross-domain consistency.

\section{Methods}
\subsection{NeRF and EG3D}

\noindent\textbf{NeRF} \cite{mildenhall2021nerf} offers an implicit representation of 3D scenes by employing a 5D vector-valued function. This function takes as input a 3D spatial location, denoted as $\mathbf{x} = (x, y, z)$, and a 2D viewing direction, represented as $\mathbf{d} = (\theta, \phi)$. The output of this function encompasses the emitted color, denoted as $\mathbf{c} = (r, g, b)$, as well as the volume density, symbolized by $\sigma$. In practical implementation, this continuous 5D representation is approximated through a Multi-Layer Perceptron (MLP), denoted as $F_{\Psi} : (\mathbf{x}, \mathbf{d}) \rightarrow (c, \sigma)$. Within the NeRF framework, every scene is characterized by its volume density and the radiance emitted in a particular direction. Consequently, the color of any ray traversing the scene can be rendered using principles derived from classical volume rendering \cite{kajiya1984ray}. By interpreting the volume density $\sigma (\mathbf{x})$ as the differential probability of a ray terminating at an infinitesimal particle situated at location $\mathbf{x}$, it becomes possible to calculate the expected pixel value $C(\mathbf{r})$ along a camera ray $\mathbf{r}(t) = \mathbf{o} + t\mathbf{d}$, where $\mathbf{r}(t)$ represents a ray emanating from the camera centered at position $\mathbf{o}$, with near and far bounds delineated by $t_n$ and $t_f$. This calculation is given by the integral:


\begin{equation}
\begin{aligned}
C(\mathbf{r})=\int_{t_n}^{t_f} T(t) \sigma(\mathbf{r}(t)) \mathbf{c}(\mathbf{r}(t), \mathbf{d}) d t,
\end{aligned}
\end{equation}
where $T(t) = \exp \left(-\int_{t_n}^t \sigma(\mathbf{r}(s)) ds\right)$ signifies the accumulated transmittance along the ray from $t_n$ to $t$. The trainable parameters $\Psi$ are optimized through a process of training to yield a value of $C(\mathbf{r})$ that approximates the actual pixel value observed in the ground truth data.


\noindent\textbf{EG3D} \cite{chan2022efficient} has gained considerable popularity as a 3D-aware image generation method, leveraging both GANs and NeRF. An overview of the generator architecture is presented in the top part of Figure \ref{fig:eg3d}. This architecture involves combining the latent code $\mathbf{z}$ with camera parameters $\mathbf{P}$, resulting in an intermediate latent code $\mathbf{w}$ derived via the mapping network $\mathbf{M}$. This intermediate latent code $\mathbf{w}$ is then used to modulate the Style-based Generator $\mathbf{G1}$ \cite{karras2020analyzing}, leading to the generation of feature maps $\mathbf{F}$ with dimensions $H_f \times H_f \times 3 M_f$.
The feature maps $\mathbf{F}$ undergoes a channel-wise splitting and reshaping process, yielding three $M_f$-channel planes characterized by a resolution of $H_f\times H_f\times M_f$. Furthermore, EG3D is capable of querying any 3D position $x\in \mathbb {R}^3$ by projecting it onto each of the three feature planes. Subsequently, the corresponding feature vector is retrieved through interpolation, and these three feature vectors are aggregated via summation. Additionally, an additional MLP, denoted as $\textbf{Tri-D}$, featuring a single hidden layer comprising 64 units with softplus activation functions, interprets the aggregated 3D features $\mathbf{F}$ as color $\mathbf{c}$ and density $\sigma$.

Both of these quantities are then subjected to processing by a neural volume renderer \cite{max1995optical}, facilitating the projection of the 3D feature volume into a 2D feature image. It's worth noting that, unlike the volume rendering approach in NeRF \cite{mildenhall2021nerf}, EG3D's volume rendering produces feature images, specifically a 32-channel feature image $\mathbf{I_F}$, as opposed to RGB images. This choice is made in recognition of the fact that feature images inherently contain a richer information content that can be effectively harnessed for further refinement and generation processes. Finally, the latent code $\mathbf{w}$ is also utilized to modulate the Style-based Super-resolution module $\mathbf{G2}$. This module performs the critical tasks of upsampling and refining the 32-channel feature image $\mathbf{I_F}$, yielding the final RGB image $\mathbf{I_{rgb}}$ with dimensions $H\times W\times 3$.

\subsection{Training Parameters Determination}

In our approach, we adapt a pre-trained EG3D generator \cite{chan2022efficient} using information from a single reference image. Notably, when faced with limited training data, 3D GANs tend to experience significant performance degradation compared to their 2D counterparts \cite{yang2022improving}. Consequently, our methodology involves identifying a subset of model parameters that offer the requisite expressiveness for downstream adaptation. To determine the most effective components for fine-tuning, we conduct an extensive series of ablation studies on various module elements within EG3D. These elements include the Mapping Network ($\mathbf{M}$), the Style-based Generator ($\mathbf{G1}$), the Tri-plane Decoder ($\textbf{Tri-D}$), and the Style-based Super-resolution Module ($\mathbf{G2}$).


To achieve adaptation to the target domain using only a single reference image, we employ the original adversarial loss to fine-tune different components of the original EG3D generator. As depicted in Figure \ref{fig:ablation_module}, we compare the original images generated by the pre-trained generator with the results of five distinct adaptation approaches. We select a sketch image ($I_{tar}$ in Figure \ref{fig:eg3d}) as the target domain and follow the exact training settings of EG3D \cite{chan2022efficient}. The illustrated images are generated after the generator has processed the target image 10,000 times. Notably, fine-tuning only the Mapping Network ($\textbf{M Only}$) fails to achieve the necessary levels of high fidelity and cross-domain consistency. On the other hand, fine-tuning only the Style-based Generator ($\textbf{G1 Only}$) or all components of EG3D ($\textbf{All}$) results in unstable training and significant performance degradation in terms of generative quality. In contrast, focusing solely on fine-tuning the Tri-plane Decoder ($\textbf{Tri-D Only}$) or the Style-based Super-resolution Module ($\textbf{G2 Only}$) may not fully adapt to the target domain but does enable stable training and preserves cross-domain consistency. This observation suggests the potential for designing efficient training algorithms to achieve one-shot 3D GDA. Consequently, we introduce a novel training strategy that focuses on fine-tuning either the Tri-plane Decoder or the Style-based Super-resolution Module in this study. This approach ensures stable training and maintains high fidelity and cross-domain consistency.



\subsection{Loss Functions}
\label{sec:loss}

As depicted in Figure \ref{fig:ablation_module}, it becomes evident that simply relying on the adversarial loss does not effectively facilitate the adaptation of the source model to the target domain. This divergence from the principles of GDA in the 2D domain \cite{wu2023domain} underscores the unique challenges faced in the one-shot 3D GDA context. To address the issues of training instability in this scenario, we introduce four loss functions: \textit{Domain Direction Regularization}, \textit{Target Distribution Learning}, \textit{Image-Level Source Structure Maintenance}, and \textit{Feature-Level Source Structure Maintenance}, as detailed in this section. A comprehensive overview of EG3D and our proposed methodology are provided in the top and bottom parts of Figure \ref{fig:eg3d}, respectively. Specifically, the EG3D generator comprises a Synthesis Network ($G$) and a Super-Resolution Network ($S$). The Synthesis Network includes the Mapping Network (M), Style-based Generator (G1), Tri-plane Decoder, and Volume Rendering, while the Super-Resolution Network consists of the Style-based Super-resolution Module (G2). 

\noindent\textbf{Domain Direction Regularization.} 
Recent studies have provided sufficient evidence of the effectiveness of leveraging pre-trained CLIP models \cite{radford2021learning} for the purpose of transferring a source generator to target domains. This applicability extends to both zero-shot \cite{gal2021stylegan, zhang2022towards} and one-shot scenarios \cite{zhang2022towards, kwon2023one, zhu2021mind}. When compared to traditional adversarial-based methods \cite{wu2023domain, ojha2021few, zhao2022closer, xiao2022few}, CLIP-based methodologies exhibit a noteworthy advantage in terms of training stability, making them well-suited for our one-shot 3D GDA. 
In our approach, we introduce Domain Direction Regularization, leveraging the pre-trained CLIP image encoder to identify the CLIP-space direction between the source and target domains. Given the 3D source generator pre-trained on source domain (domain $A$) and a reference image $I_{tar}$ from the target domain (domain $B$), the CLIP-space domain direction between these two domains is computed as follows:

\begin{equation}
\Delta v_{dom}=v_{tar}-v_{sou},
\label{eq:image_direction}
\end{equation}
where $v_{tar}=E_I(I_{tar})$ denotes the embedding of the target domain $B$, and $E_I$ is the CLIP image encoder. $v_{sou}$ is the mean embedding of source images, \textit{i.e.}, $v_{sou}=v_{\bar{A}}=\mathbb{E}_{\boldsymbol{z} \sim \mathcal{N}(0, I)}\left[E_I\left(S_A(G_A(\boldsymbol{z}))\right)\right]$. To adapt the source generator $G_A$ to target domain $B$, we finetune the target generator by aligning the sample-based direction $\Delta v_{samp}$ with the  CLIP-space domain direction $\Delta v_{dom}$:

\begin{equation}
\mathcal{L}_{dir}=1-\frac{\Delta v_{samp} \cdot \Delta v_{dom}}{\left|\Delta {v}_{samp}\right|\left|\Delta v_{dom}\right|},
\end{equation}

\begin{equation}
\Delta v_{samp}=v_{B}-v_{A},
\label{eq:sample_direction}
\end{equation}
where $v_B=E_I(S_B(G_B(z)))$ and $v_A=E_I(S_A(G_A(z)))$. $L_{dir}$ not only compels the target generator to assimilate the knowledge specific to the target domain but also prevents it from merely duplicating the reference image. The effectiveness of $L_{dir}$ becomes evident when the domain gap between the source and target domains is relatively small. However, in scenarios where the domain gap is extensive, such as the transition from FFHQ to Sketch domains, relying solely on $L_{dir}$ often proves insufficient for achieving the desired level of high fidelity.

\noindent\textbf{Target distribution learning.} 
In order to further enhance the adaptability of the generator to the target domain and capture the domain-specific characteristics inherent in $I_{tar}$, we introduce the concept of the Relaxed Earth Movers Distance (REMD) \cite{kolkin2019style}. This is employed to achieve the objective of target distribution learning, denoted as $L_{dis}$, which is a widely adopted technique in the realm of 2D GDA \cite{zhang2022towards, zhanggeneralized}. Our approach begins by extracting the intermediate tokens from images $I_B$ and $I_{tar}$ using the k-th (k=3 if there is no additional explanation) layer of the CLIP image encoder. These extracted tokens are designated as $T_B=\{T^1_B, \cdots, T^n_B\}$ and $T_{tar}=\{T^1_{tar}, \cdots, T^m_{tar}\}$, respectively. In accordance with the methodologies outlined in prior works, such as \cite{zhang2022towards} and \cite{kolkin2019style}, we define the Relaxed Earth Movers Distance (REMD) as the metric for quantifying the disparity between the adapted distribution and the target distribution. This is formally expressed as:

\begin{equation}
L_{dis}=\max \left(\frac{1}{n} \sum_i \min _j M^{{i,j}}_{C}, \frac{1}{m} \sum_j \min _i M^{{i,j}}_{C}\right),
\end{equation}
where $M_C$ is the Cross-correlation Map in Fig. \ref{fig:eg3d}, which measures the token-wise cosince distances ($D_{\cos }(\cdot)$) form $T_B$ to $T_{tar}$. Each element of $M_C$ is computed as:

\begin{equation}
M^{i,j}_{C}=D_{\cos }\left(T^i_B, T^j_{tar}\right)=1-\frac{ T^i_B \cdot T^j_{tar}}{\|T^i_B\|\cdot\|T^j_{tar}\|}.
\end{equation}

\noindent\textbf{Image-level source structure Maintenance.}
The ability to generate diverse, cross-domain consistent, and multi-view consistent target domain images is crucial for one-shot 3D GDA. The pre-trained source generator demonstrates remarkable performance in terms of diversity and multi-view consistency. 
Consequently, the aforementioned properties can be realized by maintaining the consistency of domain-independent attributes between the adapted image and its corresponding source image. In this section, we introduce the concept of image-level source structure maintenance, denoted as $L_{I-str}$, which serves to align the domain-independent attributes between the adapted image $I_B=S_B(G_B(z))$ and its corresponding source image $I_A=S_A(G_A(z))$. Additionally, we extract the intermediate tokens from $I_B$ and $I_{A}$ using the k-th (k=3 if there is no additional explanation) layer of the CLIP image encoder. These extracted tokens are denoted as $T_B=\{T^1_B, \cdots, T^n_B\}$ and $T_{A}=\{T^1_{A}, \cdots, T^n_{A}\}$, respectively. Motivated by the insight that robust pattern recognition can be established by leveraging local self-similarity descriptors \cite{shechtman2007matching}, we align the self-correlation map between $T_B$ and $T_A$. In summary, the image-level source structure maintenance $L_{I-str}$ is defined as:

\begin{equation}
L_{I-str}=||M_A-M_B||_2,
\end{equation}
where $M_B$ and $M_A$ are self-correlation map of $T_B$ and $T_A$, respectively. 
Each element of $M_B$ and $M_A$ is computed as:
\begin{equation}
\begin{aligned}
M^{i,j}_{B}=\frac{ T^i_B \cdot T^j_{B}}{\|T^i_B\|\cdot\|T^j_{B}\|},\quad
M^{i,j}_{A}=\frac{ T^i_A \cdot T^j_{A}}{\|T^i_A\|\cdot\|T^j_{A}\|}.
\end{aligned}
\end{equation}
\begin{figure}[t!]

	\centering
	\includegraphics[scale=0.55]{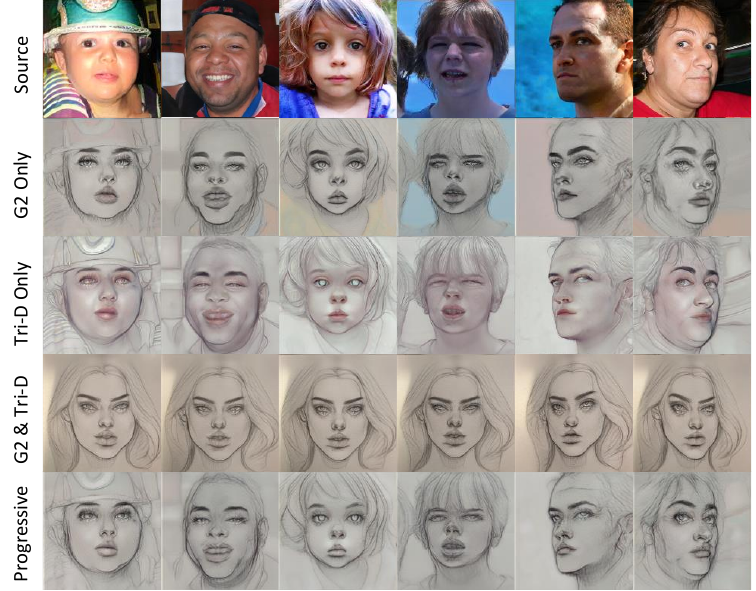}
	\caption{\textbf{Training strategy determination}. Ablation study on different fine-tuning strategies.}
	\label{fig:ablation_strategy}
\end{figure}

\noindent\textbf{Feature-level source structure Maintenance.}
In addition, we introduce a feature-level source structure maintenance term, denoted as $L_{F-str}$, to align the domain-independent attributes between the adapted image $I_B$ and its corresponding source image $I_A$. As depicted in Figure \ref{fig:eg3d}, the EG3D generator is composed of two essential parts: $G(\cdot)$ and $S(\cdot)$. The feature-level source structure maintenance operation is applied to the feature image $I_F$ generated by the $G(\cdot)$ model, specifically, the output of the volume rendering process. Notably, EG3D employs volume rendering to produce feature images, as opposed to RGB images, owing to the enhanced information content that can be harnessed for image-space refinement. In our configuration, the features are rendered with 32 channels at a resolution of $64\times 64$. The primary objective of $L_{F-str}$ is to align the most similar channels between $I_{F_B}$ and $I_{F_A}$. To this end, we first define the channel-wise features $F_B$ and $F_A$ as $F_B=\{I^1_{F_B}, \ldots, I^k_{F_B}\}$ and $F_A=\{I^1_{F_A}, \ldots, I^k_{F_A}\}$, respectively. The ultimate feature-level source structure maintenance loss is defined as:

\begin{equation}
\begin{aligned}
&h_i=\min _j \left(1-\frac{ I^i_{F_B} \cdot I^j_{F_A}}{\|I^i_{F_B}\|\cdot\|I^j_{F_A}\|} \right), H=\{h_1,\cdots,h_k\}\\
&L_{F-str}=\frac{1}{t}\sum_i\mathbb{I}\big(h_i \in \text{Minimum}(H,t)\big)\cdot h_i,
\label{eq:t}
\end{aligned}
\end{equation}
where $\mathbb I(\cdot)$ is the indicator function, and $\text{Minimum}(H,t)$ function represents the smallest $t$ elements in the list $H$. In our settings, we set $t=20$ to ease restrictions on consistency and avoid generators not being well adapted to the target domain.


\subsection{Progressive Fine-tuning Strategy}

While our proposed loss functions and determined training sub-parameters significantly alleviate training instability in one-shot 3D GDA, it is essential to address the issue of under-fitting, as evident when directly fine-tuning the networks (illustrated in Figure \ref{fig:ablation_strategy} \footnote{ In our ablation study, we examine different fine-tuning strategies using a sketch image ($I_{tar}$ in Fig. \ref{fig:eg3d}) as the target domain. These strategies incorporate the four loss functions detailed in Section \ref{sec:loss}. The images presented in the results are generated once the generator has been trained on 1000 iterations. It is important to note that the images obtained from the fine-tuning of both the Tri-plane Decoder and Style-based Super-resolution Module (G2 \& Tri-D) are produced after the generator has been trained on 500 iterations.}), where both fine-tuning of the Tri-plane Decoder (Tri-D Only) and Style-based Super-resolution Module (G2 Only) results in under-fitting, while fine-tuning both of these components (G2 \& Tri-D) leads to severe over-fitting. To address this challenge, we introduce a two-step progressive fine-tuning strategy. 

\textbf{Step 1:} Fine-tuning the Tri-plane Decoder (Tri-D) with the following objective functions:

\begin{equation}
\begin{aligned}
    \hat\theta_{Tri-D} =\arg\min_{\theta_{Tri-D}}\lambda_{dir}L_{dir}+\lambda_{dis}L_{dis}\\
    +\lambda_{I-str}L_{I-str}+\lambda_{F-str}L_{F-str}.
    \end{aligned}
\end{equation}

\textbf{Step 2:} Fine-tuning the Style-based Super-resolution Module (G2) with the following objective functions:

\begin{equation}
\begin{aligned}
    \hat\theta_{G2} =\arg\min_{\theta_{G2}}\lambda_{dir}L_{dir}+\lambda_{dis}L_{dis}
    +\lambda_{I-str}L_{I-str}.
    \end{aligned}
\end{equation}

It is important to note that $L_{F-str}$ has no impact during the fine-tuning of the G2, and therefore, it is omitted in \textbf{Step 2}. In our experiments, we use $\lambda_{dir}=1$, $\lambda_{dis}=2$, $\lambda_{I-str}=3$, and $\lambda_{F-str}=5$. As depicted in Figure \ref{fig:ablation_strategy}, our proposed progressive fine-tuning strategy demonstrates superior performance compared to other single-step training approaches.
\section{Experiments}
This section is dedicated to empirical validation of the advancements introduced by our 3D-Adapter. We begin by outlining the experimental settings in Section \ref{sec:Settings}, subsequently presenting both quantitative and qualitative results in Section \ref{sec:main experiments}. In Section \ref{sec:user_study}, we provide insights from a user study. Ablation studies are conducted to assess the impact of individual components in Section \ref{sec:ablation}. Moreover, Section \ref{sec:zero-shot} demonstrates the seamless extension of our 3D-Adapter to zero-shot GDA scenarios, yielding impressive outcomes. Section \ref{sec:extension} encompasses additional results showcasing latent space interpolation, inversion, and editing capabilities on the adapted generator. Finally, Section A of the Appendix uploads a set of videos showcasing both one-shot and zero-shot GDA.

\subsection{Experimental Settings}
\label{sec:Settings}

\noindent\textbf{Implementation.} 
Building upon prior research \cite{abdal20233davatargan}, our foundation rests on the EG3D generator \cite{chan2022efficient}, which has been pre-trained on the FFHQ dataset \cite{karras2019style}. Furthermore, we leverage the pre-trained CLIP model \cite{radford2021learning}, specifically ViT-B/16 and ViT-B/32, to implement our domain direction regularization, target distribution learning, and image-level source structure maintenance. To capture the desired image features effectively, we extract information from the third layer of the CLIP image encoder. In the domain direction regularization, we extract CLIP image features from a corpus of 5000 source images to derive the mean image embedding of the source domain, represented as $v_{sou}$. Our training employs the ADAM optimizer, with a learning rate set at 0.0025, and a batch size of 16. Fine-tuning the generator encompasses approximately 600 iterations in training Step 1, followed by an additional 1200 iterations in training Step 2. 
\begin{figure*}[t!]
	\centering
	\includegraphics[scale=0.38]{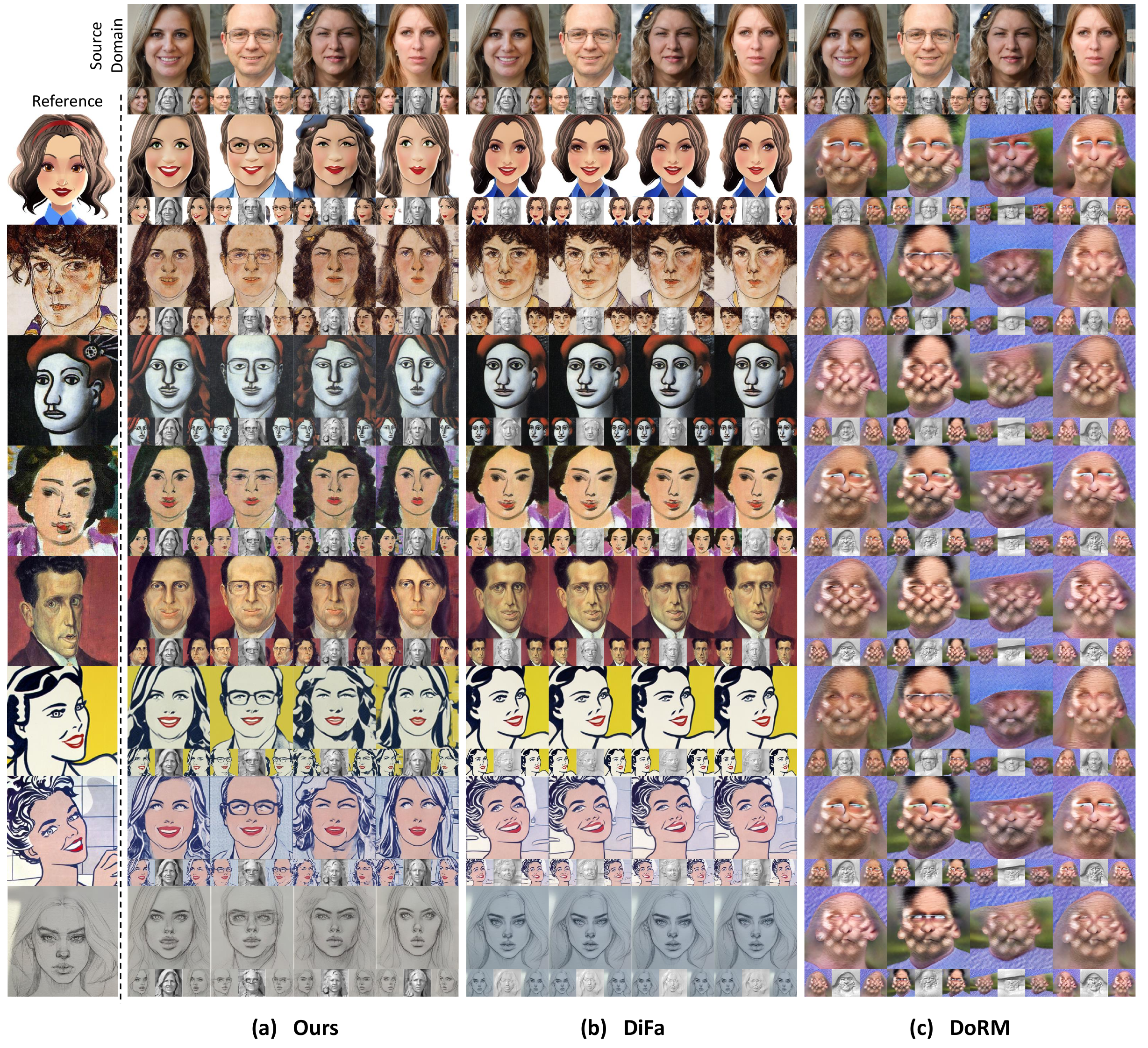}
	\caption{\textbf{Qualitative comparisons} on one-shot setting between our proposed method, DiFa \cite{zhang2022towards}, and DoRM \cite{wu2023domain}. The first row and first column show different images in source domains and reference images in target domains. \textbf{Results best seen at 500\% zoom}.}
	\label{fig:one-shot}
\end{figure*}

\noindent\textbf{Datasets.} 
Consistent with prior work on domain adaptation for 3D GANs \cite{abdal20233davatargan}, the FFHQ dataset \cite{karras2019style}, featuring images with a resolution of $512\times512$, is designated as the source domain for the experimental framework presented in this paper. In our one-shot evaluation, we employ three datasets for quantitative assessment: Cartoons \cite{yang2022pastiche}, Sketches \cite{wang2008face}, and Ukiyoe \cite{pinkney2020resolution}, comprising approximately 300, 300, and 5200 images, respectively. Furthermore, we expand our analysis to include various reference images as different target domains, and we illustrate the corresponding qualitative results in Figure \ref{fig:one-shot}.
\begin{table}[htbp]
\footnotesize
\tiny
\centering
\tabcolsep=0.17cm
\caption{\textbf{Quantitative evaluation on one-shot GDA}. Evaluation metrics include FID ($\downarrow$), KID ($\downarrow$), ID ($\uparrow$), Depth ($\downarrow$), Intra-ID ($\uparrow$), LIQE ($\uparrow$), and Q-Align ($\uparrow$). }
\begin{tabular}{r c cc cc cc}
\hline
Datasets & \multicolumn{5}{c}{Cartoon}\\
\hline
Method & FID &KID& ID& Depth&Intra-ID&LIQE&Q-Align\\
\hline 
DiFa  &175.7&0.193&0.227&0.081&\textbf{0.928}&\textbf{0.792}&\textbf{0.819}\\
DoRM  &244.6&0.288&0.151&0.136&0.909&0.091&0.075\\
\hline
\textbf{Ours} &\textbf{132.6}&\textbf{0.101}&\textbf{0.463}&\textbf{0.014}&0.913&{0.789}&0.813 \\
\hline
\end{tabular}
\begin{tabular}{rc c ccccc}
\hline
Datasets&\multicolumn{5}{c}{Sketches}\\
\hline
Method &FID &KID& ID& Depth&Intra-ID&LIQE&Q-Align\\
\hline 
DiFa  &219.4&0.361&0.329&0.055&\textbf{0.936}&0.661&{0.675}\\
DoRM  &382.1&0.540&0.070&0.117&0.913&0.088&0.071\\
\hline
\textbf{Ours} &\textbf{59.59}&\textbf{0.067}&\textbf{0.428}&\textbf{0.013}&0.922&\textbf{0.673}&\textbf{0.681}\\
\hline
\end{tabular}
\begin{tabular}{r c cc cccc}
\hline
Datasets&\multicolumn{5}{c}{Ukiyoe}\\
\hline
Method &FID &KID& ID& Depth&Intra-ID&LIQE&Q-Align\\
\hline
DiFa  &197.3&0.259&0.089&0.100&\textbf{0.966}&{0.659}&{0.675}\\
DoRM  &323.1&0.346&0.045&0.175&0.924&0.090&0.074\\
\hline
\textbf{Ours} &\textbf{118.7}&\textbf{0.186}&\textbf{0.365}&\textbf{0.013}&0.937&\textbf{0.664}&\textbf{0.680}\\
\hline
\end{tabular}
\label{table:1-shot}
\end{table}

\noindent\textbf{Metrics.} 
Consistent with established practices in 2D-GDA \cite{ojha2021few,zhao2022closer,zhang2022towards}, we employ two key metrics to assess the quality of synthesis, namely the Fréchet Inception Distance (FID) \cite{heusel2017gans} and the Kernel Inception Distance (KID) \cite{binkowski2018demystifying}. Both FID and KID are indicative of synthesis quality, where lower values are preferable, and are used to simultaneously measure high fidelity and diversity. The FID and KID scores are computed by comparing 5000 generated images with the entire set of images in the target dataset. In addition, we employ the Identity (ID) similarity metric \cite{zhanggeneralized}, as determined by Arcface \cite{deng2019arcface}, to assess the extent to which identity information from source images is preserved in the adapted images. Higher ID similarity values are indicative of better preservation of identity information and, therefore, higher cross-domain consistency. Furthermore, we introduce a depth difference metric (Depth), which quantifies the geometric consistency across domains. This is calculated by synthesizing 5000 source images and their corresponding adapted images and computing the Mean Squared Error (MSE) between their corresponding depth maps, as expressed by:

\begin{table*}[htbp]
\caption{User study on one-shot 3D GDA. The numbers represent the percentage of users who favor the images synthesized corresponding method among the all three methods.
\label{Tab:user_study}}
\setlength{\abovecaptionskip}{0.1cm}
    \setlength{\belowcaptionskip}{-0.3cm}
    \centering
	\begin{tabular}{c|c|c|c}	
\hline
Model Comparison & image quality & style similarity &attribute consistency\\
\hline 
DiFa \cite{zhang2022towards} &21.09\%&55.76\%&0\%\\
DoRM \cite{wu2023domain} &0\%&0\%&0\%\\
Ours&78.91\%&44.24\%&100\%\\
\hline
\end{tabular}
\end{table*}

\begin{equation}
\label{eq:depth}
    \text{Depth} = E_{z\sim p(z)}\|\text{D}(G_s(z))-\text{D} (G_t(z))\|,
\end{equation}
where $\text{D}(\cdot)$ represents the generated depth map, and $G_s$ and $G_t$ denote the source domain generator and the target domain generator, respectively. We also assess multi-view consistency by evaluating the cosine similarity of facial identities, using the ArcFace \cite{deng2019arcface} metric. To perform this evaluation, we generate 5000 random faces and render two views for each face, each from poses randomly selected from the source dataset pose distribution. The facial identity similarity, expressed as the mean score, is referred to as intra-identity similarity (Intra-ID), where higher values are indicative of better multi-view consistency. Finally, we adopt two popular no-reference quality metrics LIQE \cite{zhang2023blind} and  Q-Align \cite{wu2023q} to compare the generated image quality.

\subsection{Quantitative and Qualitative Results}
\label{sec:main experiments}
\noindent\textbf{Qualitative comparison.} 
In Figure \ref{fig:one-shot}, we present qualitative comparisons using the FFHQ dataset as the source domain. These results highlight the challenges and nuances encountered in the context of one-shot 3D GDA. Firstly, we consider the adversarial-based approach "DoRM" \cite{wu2023domain}. Our observations indicate that it exhibits significant issues, including unstable training dynamics and substantial performance degradation. The resulting images often fail to meet the desired standards of one-shot 3D GDA, reflecting the inherent difficulties in employing adversarial loss for this task. We then turn our attention to the non-adversarial-based DiFa method \cite{zhang2022towards}. DiFa operates by fine-tuning the entire generator, which results in a severe form of model collapse. A notable manifestation of this is the generation of nearly identical images, as illustrated in Figure \ref{fig:one-shot} (b), where the generated images closely resemble the reference images. This aspect highlights a significant limitation in DiFa's performance, particularly in terms of diversity and adaptability. Additionally, we have incorporated comparison results with other baselines, including One-shot CLIP \cite{kwon2023one}, Few-shot GAN adaptation \cite{ojha2021few}, and Mind the Gap \cite{zhu2021mind}, which are detailed in Section D of the Appendix. Furthermore, we provide qualitative results for one-shot 3D GDA on the AFHQ-Cat domain and cross-domain adaptation in Section B and Section C of the Appendix, respectively.


In contrast, our proposed approach demonstrates a superior capacity to capture domain-specific characteristics from a single reference image while retaining substantial information related to the identity and structure of the source image. Our method surpasses existing approaches by effectively addressing the dual challenges of domain adaptation and identity preservation.

\noindent\textbf{Quantitative comparison.} 
We also conducted a quantitative comparative analysis between our proposed method and other existing methodologies \cite{zhang2022towards,wu2023domain}. This evaluation was performed under three experimental configurations, namely, FFHQ $\rightarrow$ {Cartoon, Sketches, Ukiyoe}. For each of these settings, we adopted a randomized approach, selecting one image from the respective target dataset for adaptation. The results encompassing all seven critical metrics are presented in Table \ref{table:1-shot}. To mitigate potential random sampling variability, we conducted five iterations of the adaptation process and computed the mean values as the final scores. Our analysis showcases the superior performance of our proposed method in comparison to the baseline approaches. Specifically, in comparison to state-of-the-art methods within the realm of 2D One-shot GDA, our proposed method exhibits significant improvements across all desired attributes, including target domain consistency, extensive diversity, cross-domain consistency, and multi-view consistency. It is pertinent to note that the DiFa method \cite{zhang2022towards} exhibits a pronounced susceptibility to model collapse, thereby yielding images with striking similarities across different camera poses. Consequently, DiFa excels over our proposed method in terms of Intra-ID.

\subsection{User Study}
\label{sec:user_study}
\begin{figure*}[t!]
	\centering
	\includegraphics[scale=0.23]{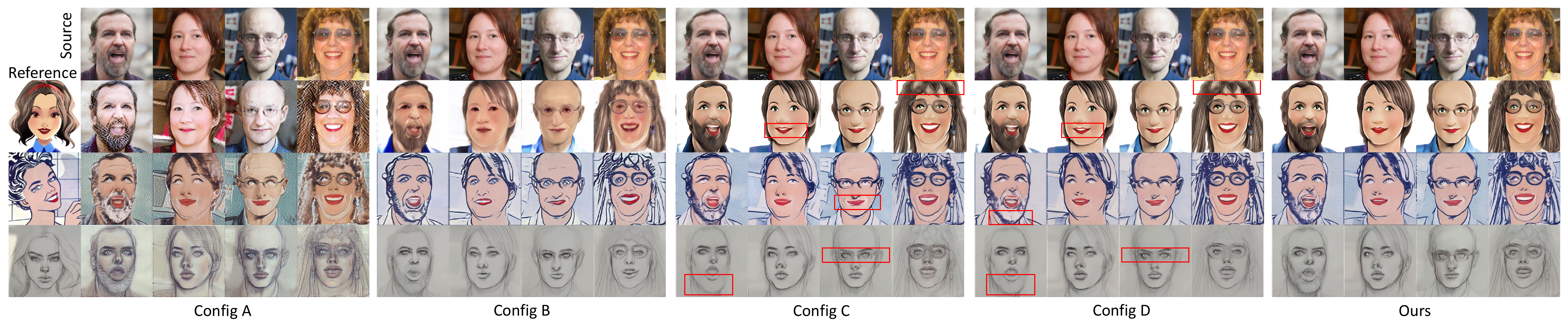}
	\caption{\textbf{Ablation study of different training losses.}  \textcolor{red}{Red} boxes indicate the difference between adaptive target images and corresponding source images. \textbf{Results best seen at 500\% zoom}.}
	\label{fig:ablation_loss}
\end{figure*}

To provide comprehensive evaluation of our approach, we conducted a user study designed to complement quantitative metrics. This user study, involving feedback from participants, was instrumental in providing a holistic understanding of our method's performance. Specifically, we presented users with a set of reference, source, and three adapted images generated by various methods. We tasked users with selecting the most suitable adapted image based on three criteria: image quality, style similarity with the reference, and attribute consistency with the source image. To ensure the robustness of our findings, we generated a substantial sample size of 500 images for each method and enlisted the feedback of 50 users. Each user was randomly assigned 50 samples and given ample time to complete the task. The results, as depicted in Table \ref{Tab:user_study}, demonstrate a strong preference for our method across all three evaluation aspects, with particularly notable favorability in terms of image quality and attribute consistency. It is noteworthy that the DiFa method \cite{zhang2022towards} is susceptible to severe mode collapse, often leading to the replication of reference images. Consequently, it garners favor on style similarity but lags behind in other critical aspects of evaluation.

\begin{table}[htbp]
\footnotesize
\centering
\caption{\textbf{Ablation study of different training strategies in quantitative evaluation.}  Noting that G2 \& Tri-D suffers from severe model
collapse and generates similar images with the different
camera poses. Therefore, G2 \& Tri-D method has the better Intra-ID
than our proposed method.}
\begin{tabular}{r c cc cc}
\hline
Datasets & \multicolumn{5}{c}{Cartoon}\\
\hline
Method & FID &KID& ID& Depth&Intra-ID\\
\hline 
Tri-D Only  &141.6&0.121&0.427&0.026&0.906\\
G2 Only  &149.1&0.132&0.409&0.028&0.901\\
G2 \& Tri-D &171.2&0.190&0.230&0.079&\textbf{0.929}\\
\textbf{Ours} &\textbf{132.6}&\textbf{0.101}&\textbf{0.463}&\textbf{0.014}&0.913 \\
\hline
\end{tabular}
\begin{tabular}{rc c ccc}
\hline
Datasets&\multicolumn{5}{c}{Sketches}\\
\hline
Method &FID &KID& ID& Depth&Intra-ID\\
\hline 
Tri-D Only & 67.3&0.083&0.412&0.022&0.908\\
G2 Only  &71.5&0.088&0.409&0.024&0.901\\
G2 \& Tri-D&215.3&0.358&0.330&0.054&\textbf{0.936}\\
\textbf{Ours} &\textbf{59.59}&\textbf{0.067}&\textbf{0.428}&\textbf{0.013}&0.922\\
\hline
\end{tabular}
\begin{tabular}{r c cc cc}
\hline
Datasets&\multicolumn{5}{c}{Ukiyoe}\\
\hline
Method &FID &KID& ID& Depth&Intra-ID\\
\hline
Tri-D Only  &126.1&0.199&0.349&0.021&0.925\\
G2 Only  &126.9&0.205&0.345&0.028&0.921\\
G2 \& Tri-D&195.2&0.261&0.087&0.101&\textbf{0.963}\\
\textbf{Ours} &\textbf{118.7}&\textbf{0.186}&\textbf{0.365}&\textbf{0.013}&0.937\\
\hline
\end{tabular}
\label{table:Ablation_training}
\end{table}
\begin{figure*}[t!]
	\centering
	\includegraphics[scale=0.33]{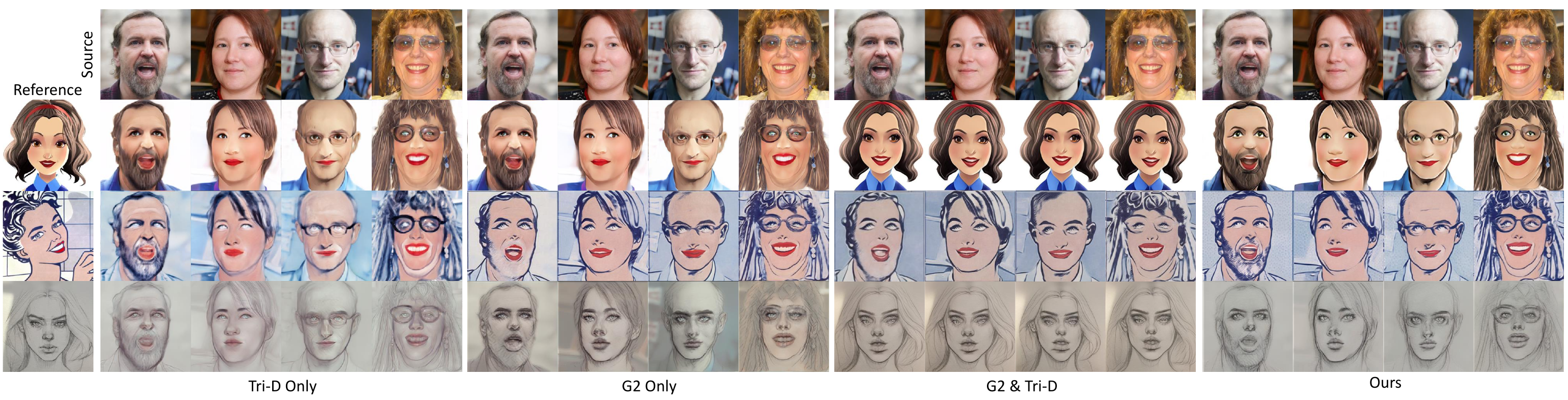}
	\caption{\textbf{Ablation study of different training strategies.}  Results best seen at 500\% zoom.}
	\label{fig:ablation_training_strategies}
\end{figure*}

\subsection{Ablation Study}
\label{sec:ablation}
Ablation studies are conducted to evaluate the effects of different critical components of our proposed method, \textit{i.e.}, the progressive training strategy and different advanced loss functions. Additionally, we explore the influence of different selections of CLIP's layers on both target distribution learning and image-level source structure maintenance.

\noindent\textbf{Ablation study of training strategies.}
Similar to Figure \ref{fig:ablation_strategy}, we extended our investigation by conducting additional ablation studies on different fine-tuning strategies. As shown in Figure \ref{fig:ablation_training_strategies} and Table \ref{table:Ablation_training}, it becomes evident that a direct approach to network fine-tuning faces challenges such as underfitting when fine-tuning either the Tri-plane Decoder (Tri-D Only) or the Super-resolution Network (G2 Only), and severe overfitting when fine-tuning both the Tri-plane Decoder and Super-resolution Network (G2 \& Tri-D). Conversely, our proposed two-step progressive fine-tuning strategy consistently demonstrates exceptional performance across diverse datasets.

\begin{figure*}[t!]
	\centering
	\includegraphics[scale=0.42]{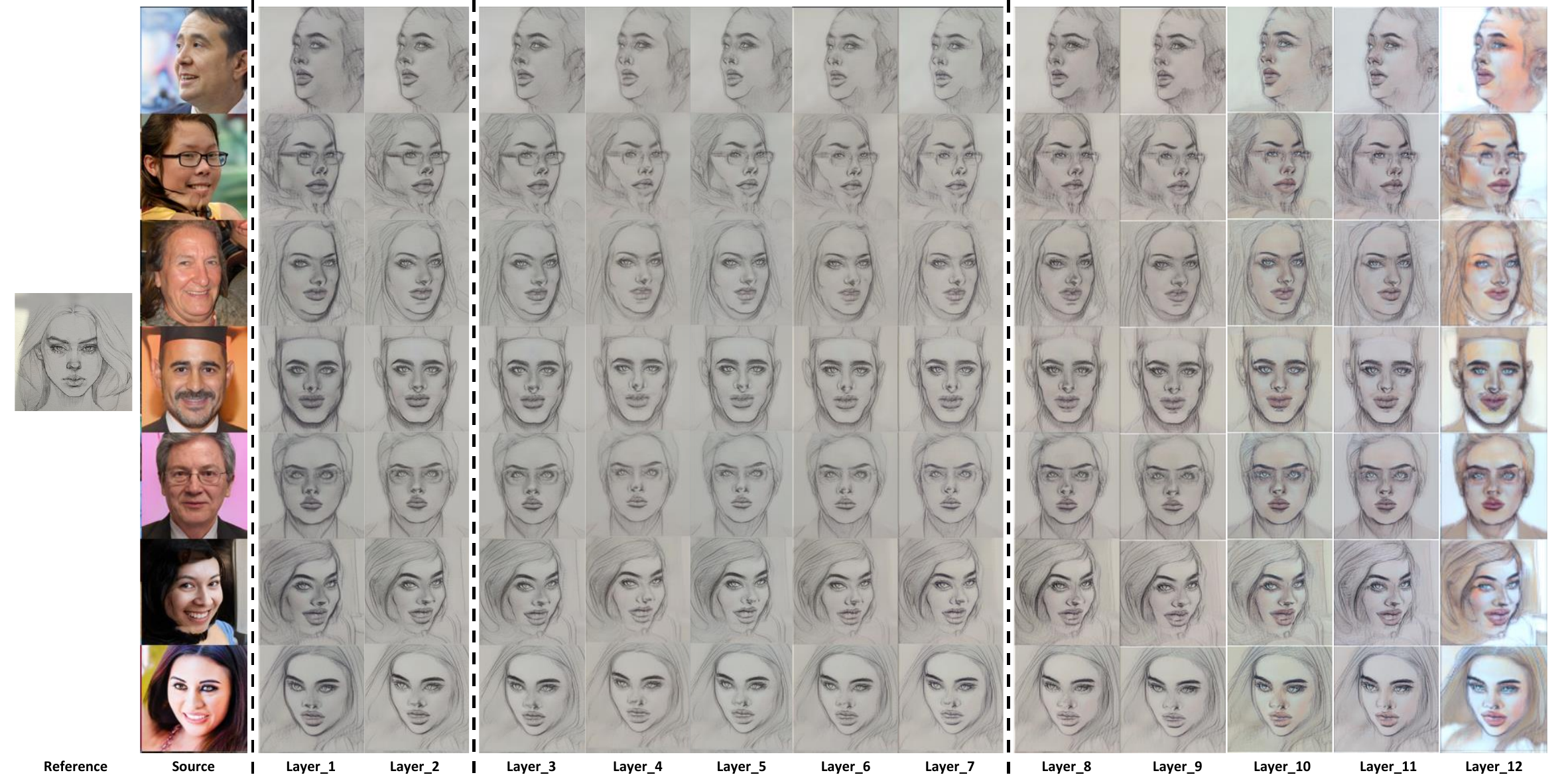}
	\caption{\textbf{Ablation studies on the layer choice of the pre-trained CLIP.} Different columns showcase the results obtained under various configurations. These configurations involve the utilization of intermediate tokens from diverse layers of the CLIP image encoder. }
	\label{fig:clip_ablation}
\end{figure*}

\begin{table}

	\caption{\textbf{The configuration for the ablation study of training losses.} The value of different hyperparameter is set to: $\lambda_{dir}=2$, $\lambda_{dis}=1$, $\lambda_{I-str}=3$, and $\lambda_{F-str}=5$ in ours, and other configurations make one loss function with a hyperparameter of 0, respectively. 
		\label{Tab:abalation study_loss}}
	\centering
	\begin{tabular}{c|ccccc}
		\toprule
		Configuration&$\lambda_{dir}$&$\lambda_{dis}$&$\lambda_{I-str}$&$\lambda_{F-str}$\\
  \hline
        $\mathbf{Config} \textbf{A} $:&\checkmark&&\checkmark&\checkmark \\
        $\mathbf{Config} \textbf{B} $: &&\checkmark&\checkmark&\checkmark\\
        $\mathbf{Config} \textbf{C} $: &\checkmark&\checkmark&&\checkmark\\
        $\mathbf{Config} \textbf{D} $: &\checkmark&\checkmark&\checkmark&\\
        $\mathbf{Ours} $:&\checkmark&\checkmark&\checkmark&\checkmark \\
		
		\bottomrule
	\end{tabular}
\end{table}

\noindent\textbf{Ablation study of training losses.}
Our proposed method comprises four loss functions. In this section, we perform an ablation study on the configurations shown in Table \ref{Tab:abalation study_loss} to demonstrate the effectiveness of the proposed method. The outcomes, depicted in Figure \ref{fig:ablation_loss} and Table \ref{table:Ablation_losses}, reveal the following key insights:
i) The results of \textbf{Config A} highlight that incorporating Target Distribution Learning ($\lambda_{dis}$) plays a pivotal role in capturing essential information about the target domain. The removal of this component results in generated images lacking the characteristic of the target domain.
ii) The results of \textbf{Config B} underscore the significance of Domain Direction Regularization ($\lambda_{dir}$), which ensures target domain consistency while enhancing diversity and cross-domain consistency.
iii) The results of \textbf{Config C} and \textbf{Config D} demonstrate that both Image-level Source Structure Maintenance ($\lambda_{I-str}$) and Feature-level Source Structure Maintenance ($\lambda_{F-str}$) yield improvements in diversity, cross-domain consistency, and multi-view consistency. The incorporation of $\lambda_{I-str}$ and $\lambda_{F-str}$ prompts the adapted generator to retain domain-sharing attributes, such as hairstyle, facial features, glasses, and mustaches. Consequently, the generator inherits diverse generation capabilities from the pre-trained model.

This comprehensive ablation study provides empirical evidence of the effectiveness of our proposed method and the crucial role each loss function plays in enhancing various aspects of the adaptation process.

\begin{table}[ht]
\footnotesize
\centering
\caption{\textbf{Ablation study of different training losses in quantitative evaluation.}  Noting that Config A cannot adapt to target domain successfully. Therefore, Config A method has the better ID and Depth than our proposed method. }
\begin{tabular}{r c cc cc}
\hline
Datasets & \multicolumn{5}{c}{Cartoon}\\
\hline
Method & FID &KID& ID& Depth&Intra-ID\\
\hline 
Config A&213.9& 0.251&\textbf{0.573} &\textbf{0.009}& \textbf{0.917}\\
Config B&148.6&0.132&0.448&0.015&0.912\\
Config C&139.2&0.107&0.458&0.021&0.910\\
Config D&138.1&0.106&0.460&0.018&0.910\\ 
\textbf{Ours} &\textbf{132.6}&\textbf{0.101}&{0.463}&{0.014}&0.913 \\
\hline
\end{tabular}
\begin{tabular}{rc c ccc}
\hline
Datasets&\multicolumn{5}{c}{Sketches}\\
\hline
Method &FID &KID& ID& Depth&Intra-ID\\
\hline 
Config A&307.6& 0.491&\textbf{0.473}&\textbf{0.008}&0.918\\
Config B&71.8&0.103&0.408&0.013&0.916\\
Config C&64.1&0.074&0.414&0.019&0.918\\
Config D&62.2&0.071&0.420&0.017&0.920\\
\textbf{Ours} &\textbf{59.6}&\textbf{0.067}&{0.428}&{0.013}&\textbf{0.922}\\
\hline
\end{tabular}
\begin{tabular}{r c cc cc}
\hline
Datasets&\multicolumn{5}{c}{Ukiyoe}\\
\hline
Method &FID &KID& ID& Depth&Intra-ID\\
\hline
Config A&165.8& 0.234&\textbf{0.391}&\textbf{0.009}&0.935\\
Config B&128.9&0.201&0.349&0.014&0.934\\
Config C&124.1&0.193&0.357&0.019&0.929\\
Config D&123.8&0.190&0.360&0.016&0.031\\
\textbf{Ours} &\textbf{118.7}&\textbf{0.186}&{0.365}&{0.013}&\textbf{0.937}\\
\hline
\end{tabular}
\label{table:Ablation_losses}
\end{table}

\begin{figure*}[t!]
	\centering
	\includegraphics[scale=0.55]{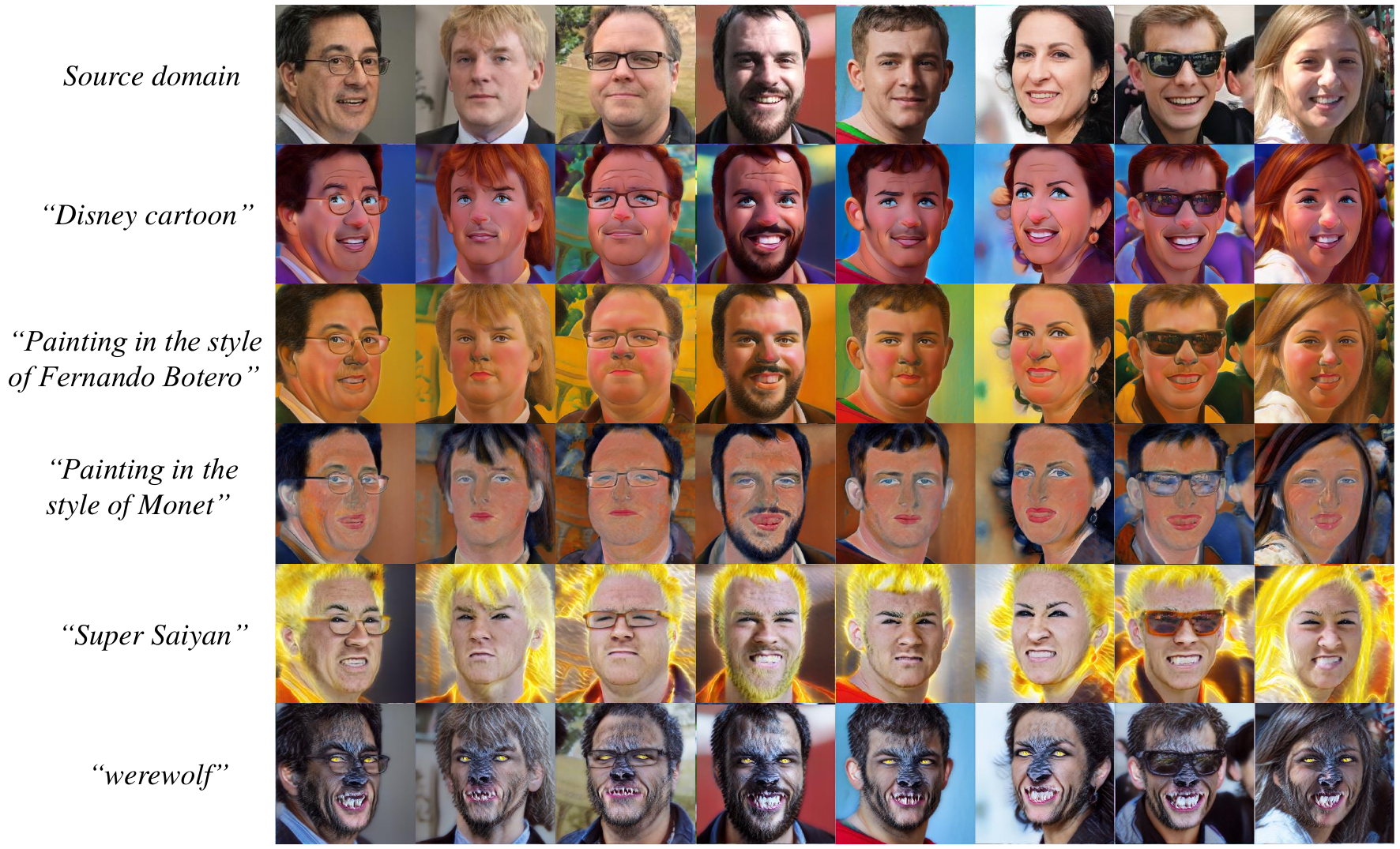}
	\caption{\textbf{Qualitative results for zero-shot 3D GDA.} The first row and first
column show the source images and the descriptions of target domain, respectively. \textbf{Results best seen at 500\% zoom}.
}
	\label{fig:zero-shot}
\end{figure*}

\noindent\textbf{Layer choice of target distribution learning and image-level source structure maintain.} 
In our primary experiments, we derived intermediate tokens from input images using the third layer (k=3) of the CLIP image encoder for both target distribution learning and image-level source structure maintenance. This section delves deeper into the exploration of different layer choices and their impact on the method's performance. It is essential to note that the CLIP image encoder functions as a Vision Transformer (ViT), comprising 12 transformer blocks and 12 hierarchical features from intermediate layers. As depicted in Figure \ref{fig:clip_ablation}, we categorize all intermediate layers into fine-level (1-2), middle-level (3-7), and coarse-level (8-12). Our observations illustrate that fine-level layers primarily capture fine-grained characteristics of the reference image and fine-grained self-correlation maps of source images, resulting in adapted images with a distinct, clear outline that diverges from the reference image. Conversely, coarse-level layers capture coarse-grained features, making the generated images less adaptable to the target domain. In contrast, middle-level layers effectively capture both representative domain styles and attributes. Thus, we default to utilizing intermediate tokens from the third layer of the CLIP image encoder, as it strikes a balance between fine-grained and coarse-grained information, facilitating the adaptation process to the target domain.


\begin{figure*}[t!]
	\centering
	\includegraphics[scale=0.6]{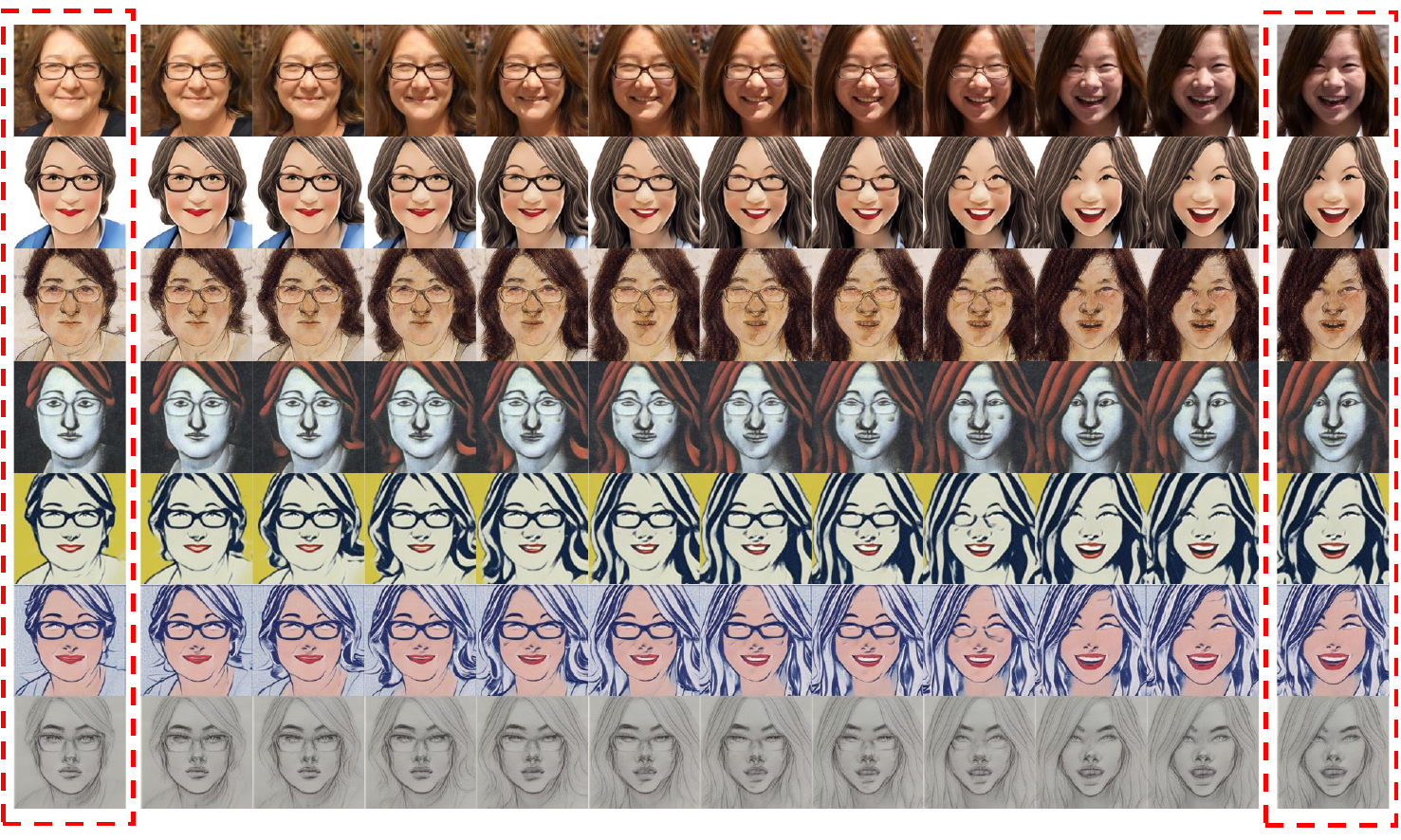}
	\caption{\textbf{Latent interpolation.} All the semantics (\textit{e.g.}, the glasses, the haircut, and the pose) vary gradually. \textbf{Results best seen at 500\% zoom}.}
	\label{fig:image_inter}
\end{figure*}

\begin{figure}[t!]
	\centering
	\includegraphics[scale=0.45]{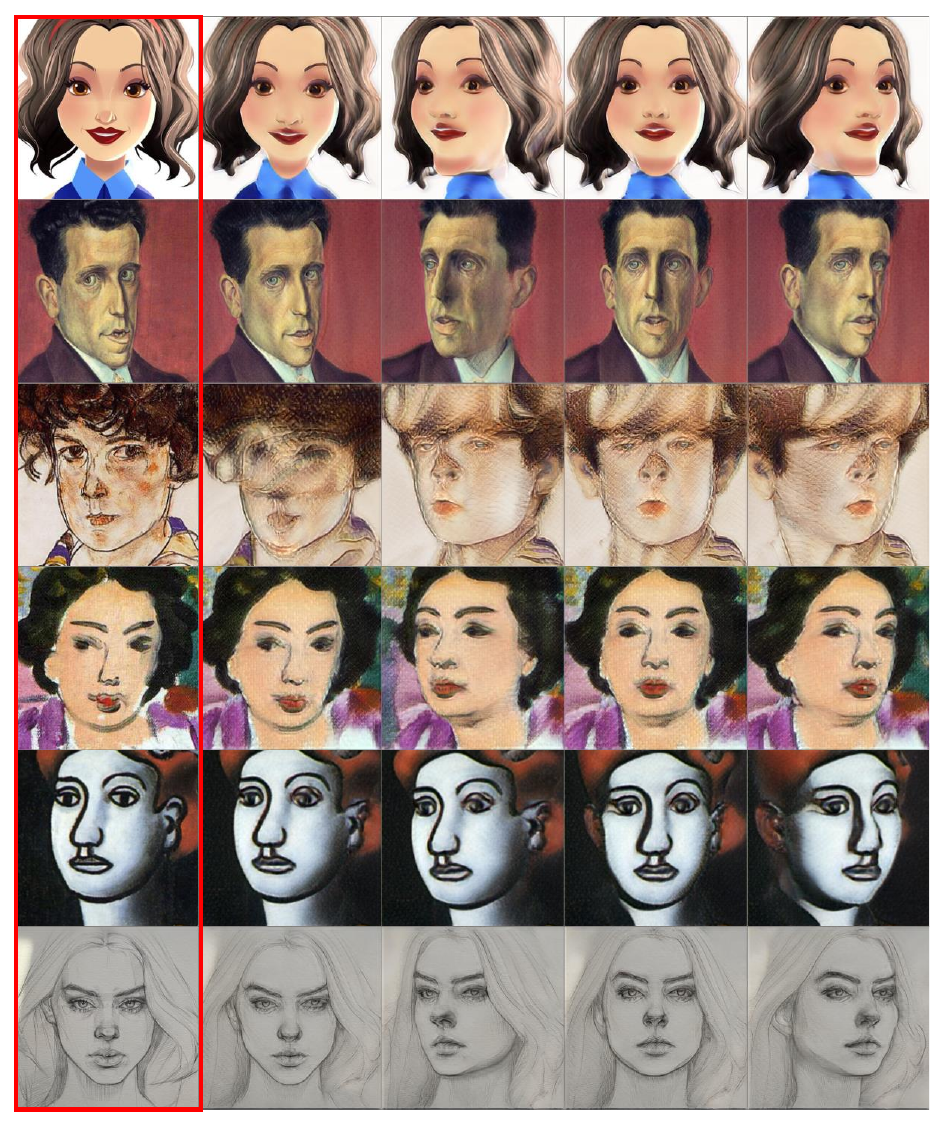}
	\caption{\textbf{3D GAN inversion.}  The initial column, denoted by the red box, contains the input image. The second column showcases the reconstructed images, while the subsequent columns display generated images featuring diverse poses. \textbf{Results best seen at 500\% zoom}.}
	\label{fig:gan_inversion}
\end{figure}
\begin{figure*}[t!]
	\centering
	\includegraphics[scale=0.42]{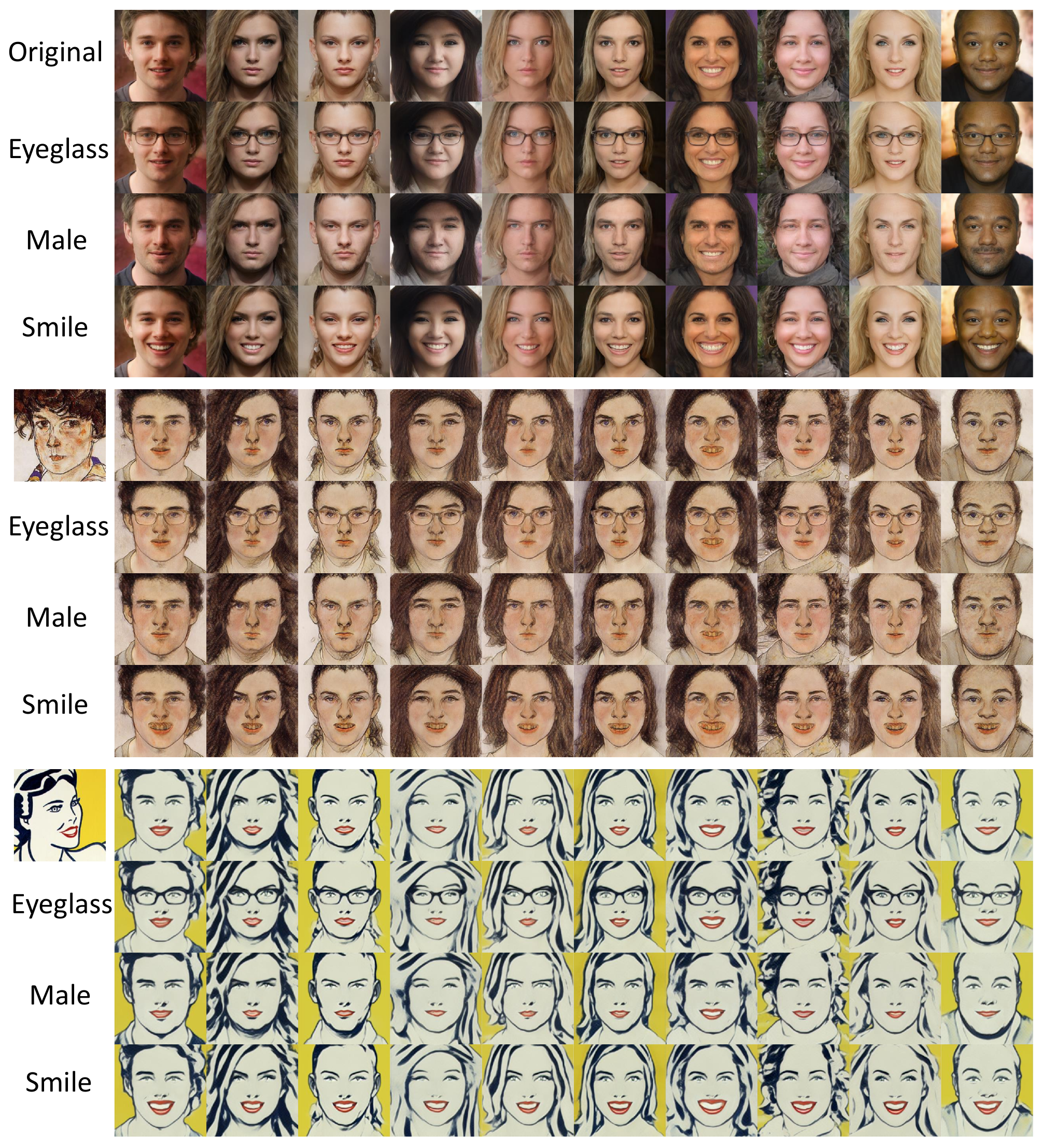}
	\caption{\textbf{Latent edit.}  All editing directions are discovered by PREIM3D \cite{li2023preim3d} in source domain. \textbf{Results best seen at 500\% zoom}.}
	\label{fig:gan_edit}
\end{figure*}
\subsection{Results on Zero-shot Generative Domain Adaption}
\label{sec:zero-shot}


While our primary focus revolves around one-shot 3D GDA, our proposed method is versatile and can also be applied to zero-shot 3D GDA scenarios. In comparison to the one-shot setting, a transition to zero-shot adaptation entails the removal of the target distribution learning loss $L_{dis}$, and a modification of the domain direction regularization loss, $L_{dir}$. To be specific, in the zero-shot configuration, we deviate from the approach outlined in Eq. \ref{eq:image_direction}, which involves computing the domain direction $\Delta v_{dom}$ utilizing the image encoder of CLIP. Instead, we compute the domain direction between the CLIP's text embedding $v_{tar\_txt}$, corresponding to the given text $T_{tar}$, and the text-based embedding $v_{sou\_txt}$ derived from the source domain. This modification accommodates the distinct requirements of zero-shot 3D GDA, expanding the applicability and flexibility of our method.

\begin{equation}
\Delta v_{dom\_txt}=v_{tar\_txt}-v_{sou\_txt},
\label{eq:text_direction}
\end{equation}
where $\Delta v_{dom\_txt}$ is the defined CLIP's text domain direction, $v_{tar\_txt}=E_T(T_{tar})$ denotes the embedding of target text $T_{tar}$, and $v_{sou\_txt}=\mathbb{E}_{\boldsymbol{t_i} \sim \mathcal{X}_T}[E_T(t_i)]$ indicates the mean embedding of $N_T$ words $\mathcal{X}_T=\{t_i\}^{N_T}_{i=1}$\footnote{$\mathcal{X}_T$=\{"person", "headshot", "participant", "face", "closeup", "filmmaker", "author","pknot","contestant", "associate", "individu", "volunteer", "michele", "artist", "director", "researcher", "cropped", "lookalike", "mozam", "ml","portrait", "organizer", "kaj", "coordinator", "appearance", "psychologist", "jha", "pupils", "subject", "entrata", "newprofile", "guterres", "staffer","diem", "cosmetic", "viewer", "assistant", "writer", "practitioner", "adolescent", "white", "elling", "nikk", "addic", "onnell", "customer", "client","simone", "greener", "candidate"\} in our experiment.}. $E_T(\cdot)$ is the CLIP text encoder. Together with the sample-based direction $\Delta v_{samp}$ computed by Eq. \ref{eq:sample_direction}, the text-based domain direction regularization is defined as:

\begin{equation}
\mathcal{L}_{dir\_text}=1-\frac{\Delta v_{samp} \cdot \Delta v_{dom\_txt}}{\left|\Delta {v}_{samp}\right|\left|\Delta v_{dom\_txt}\right|}.
\end{equation}

Consequently, the overall training pipeline in the zero-shot 3D GDA is expressed as:

\textbf{Step 1:} Fine-tuning the Tri-plane Decoder (Tri-D) with the following objective functions:

\begin{equation}
\begin{aligned}
    \hat\theta_{Tri-D} =\arg\min_{\theta_{Tri-D}}\lambda_{dir}L_{dir\_{text}}\\
    +\lambda_{I-str}L_{I-str}+\lambda_{F-str}L_{F-str}.
    \end{aligned}
\end{equation}

\textbf{Step 2:} Fine-tuning the Super-resolution Network (G2) with the following objective functions:

\begin{equation}
\begin{aligned}
    \hat\theta_{G2} =\arg\min_{\theta_{G2}}\lambda_{dir}L_{dir\_{text}}
    +\lambda_{I-str}L_{I-str}.
    \end{aligned}
\end{equation}

As shown in Figure \ref{fig:zero-shot}, we present the generative images of zero-shot 3D GDA across different target domains. Additionally, we compare our method with other state-of-the-art zero-shot GDA approaches, such as StyleGAN-Fusion \cite{song2022diffusion} and DATID-3D \cite{kim2023datid}, in the Section E of the Appendix. The results demonstrate that our proposed 3D-Adapter achieves comparable or even superior performance in all desired attributes of zero-shot GDA.


\subsection{Extensions}
\label{sec:extension}

\noindent\textbf{Results of latent interpolation.} 
We conducted latent space interpolation experiments to affirm that our proposed method does not adversely affect the acquired latent space. As depicted in Figure \ref{fig:image_inter}, the first and last columns showcase the images generated using two distinct latent codes following one-shot GDA. The intervening columns depict the outcomes achieved through linear interpolation between these two latent codes. Our results demonstrate that all intermediate images obtained through this interpolation exhibit remarkably high fidelity and cross-domain consistency. Additionally, the semantic attributes within the generated images, such as eyeglasses, hairstyle, and pose, evolve progressively throughout the interpolation process. This observation underscores that our proposed approach preserves the underlying semantic structure within the learned latent space, providing further evidence of its efficacy.


\noindent\textbf{3D GAN Inversion.}
To gain insights into the latent code capabilities of the adapted generator, we conducted GAN inversion using a well-established 3D GAN inversion technique \cite{ko20233d}. The outcomes, depicted in Figure \ref{fig:gan_inversion}, affirm that the adapted generator exhibits consistent latent code capabilities with the original generator. In the majority of instances, the inversion method \cite{ko20233d} reconstructs the provided reference image within the adapted domain, generating a multitude of images with varying poses. This analysis underscores the robustness of our adapted generator in preserving latent code attributes, ensuring its suitability for diverse generative tasks.


\noindent\textbf{Image edit.}
To explore the editing potential of real images adapted to a novel target domain, we leveraged the PREIM3D method \cite{li2023preim3d} to identify editing directions within the source domain. The outcomes, depicted in Figure \ref{fig:gan_edit}, confirm that the adapted generator retains latent-based editing capabilities comparable to the original generator. The first part shows the source images and their editing results. The remaining two parts display two popular target domain images, where the first column contains the reference target images and descriptions of edit operations. The subsequent columns present the editing operations and their corresponding results. These findings underscore the preserved editability of the adapted generator.

\section{Limitation and Conclusion}

\subsection{Limitation and Future Works}

i) Although our method can achieve appealing results in one-shot 3D GDA across different target domains, it does not always perfectly maintain cross-domain consistency in some target domains. As shown in the fifth and sixth rows of Figure \ref{fig:one-shot}, the gender of some adapted images has been altered. Therefore, more advanced loss functions should be proposed in the future to ensure the cross-domain consistency of domain-independent properties. ii) The current method can only accomplish generative domain adaptation for a single domain and is unable to retain and integrate knowledge from multiple domains. This limitation prevents adaptive generators from exploring previously unseen domains. In the future, we aim to develop a generator that can integrate multiple learned domains and synthesize hybrid domains not encountered during training.

\subsection{Conclusion}
This paper introduces 3D-Adapter, the first method for one-shot 3D GDA. 3D-Adapter contains three integral components: Firstly, we conducted an extensive investigation that revealed fine-tuning specific weight sets, Tri-D and G2, as a key strategy to enhance training stability and alleviate the challenges associated with one-shot 3D GDA. Secondly, we harnessed the power of four advanced loss functions to tackle training instability and successfully realize the four essential properties of 3D GDA. Lastly, we implemented an efficient progressive fine-tuning strategy to further augment the efficacy of our approach. Qualitative and quantitative experiments demonstrate the superiority of 3D-Adapter compared to state-of-the-art methods across a wide array of scenarios. Moreover, 3D-Adapter readily extends its capabilities to zero-shot 3D GDA, yielding compelling results. Additionally, it enables latent interpolation, image inversion, and image editing within diverse target domains. 

\section{Acknowledgments}
The work was supported by the National Natural Science Foundation of China under Grands U19B2044 and 61836011.

\bibliographystyle{spbasic}      
\bibliography{egbib}

\clearpage

\section*{Appendix}
\appendix
\section{Videos}
In this section, we have uploaded a set of videos to the supplementary materials that showcase both one-shot 3D GDA and zero-shot GDA. Specifically, we generate sixteen different videos for each target domain. These videos clearly demonstrate the effectiveness of our proposed method.

\section{Experiments on Other Source Domain}
\begin{figure*}[t!]
	\centering
	\includegraphics[scale=0.55]{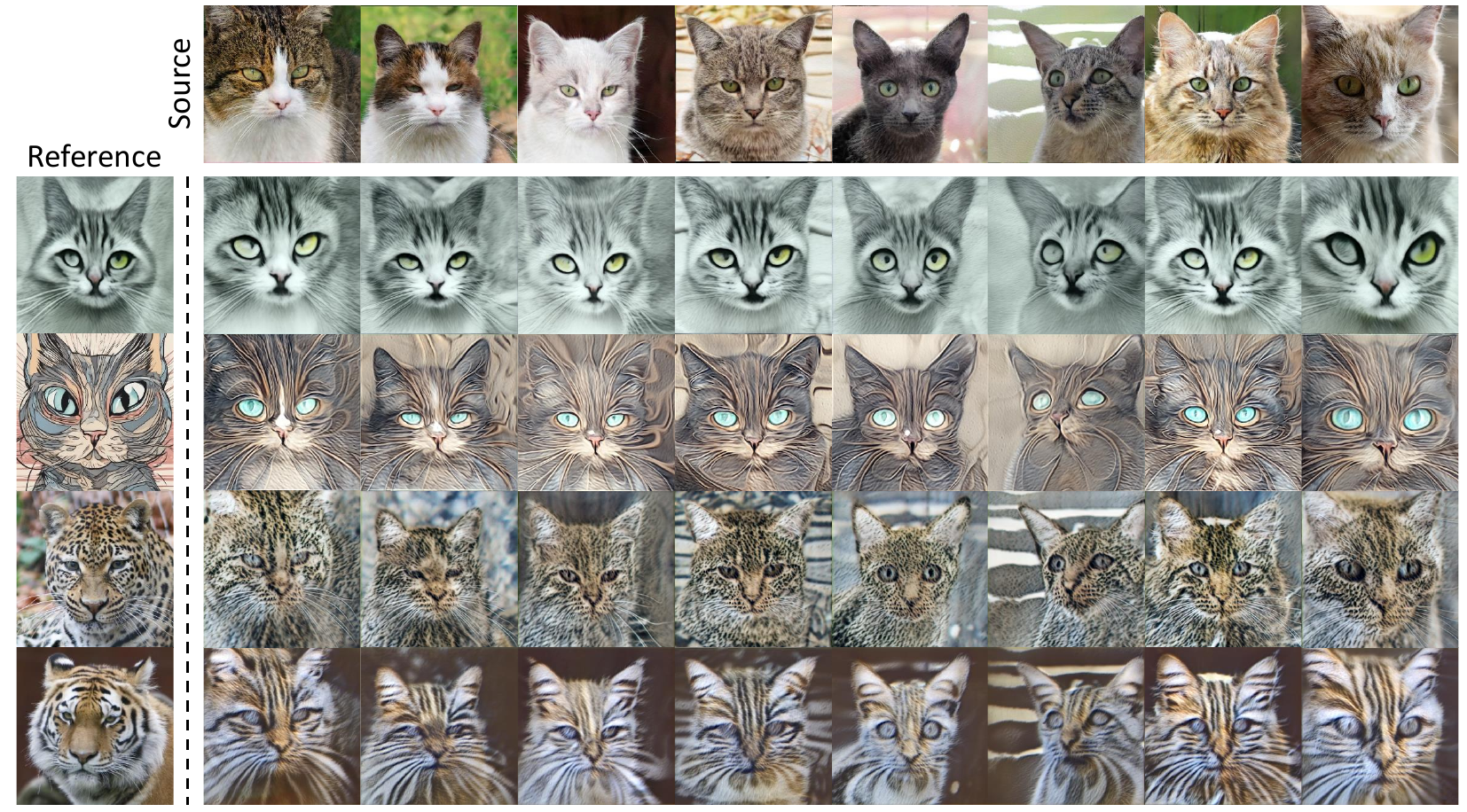}
	\caption{\textbf{Qualitative results for one-shot 3D GDA on AFHQ-Cat.} The source generator is pre-trained on AFHQ-Cat dataset. The first row and first column show the source images and the descriptions of target domain, respectively. \textbf{Results best seen at 500\% zoom}.
}
	\label{fig:one-shot-cat}
\end{figure*}

In addition to the experiments on FFHQ dataset, we conducted further one-shot 3D GDA experiments to qualitatively evaluate the effectiveness of our proposed 3D-Adapter approach. Specifically, we used the source EG3D generator pre-trained on the AFHQ-Cat dataset and adapted it to four different target domains. The results of these experiments are presented in Figure \ref{fig:one-shot-cat}. The qualitative outcomes show that our approach demonstrates a superior capacity to capture domain-specific characteristics from a single reference image while retaining substantial structural information from the source image. 

\section{Experiments on Cross-domain Adaptation}
\begin{figure*}[t!]

	\centering
	\includegraphics[scale=0.35]{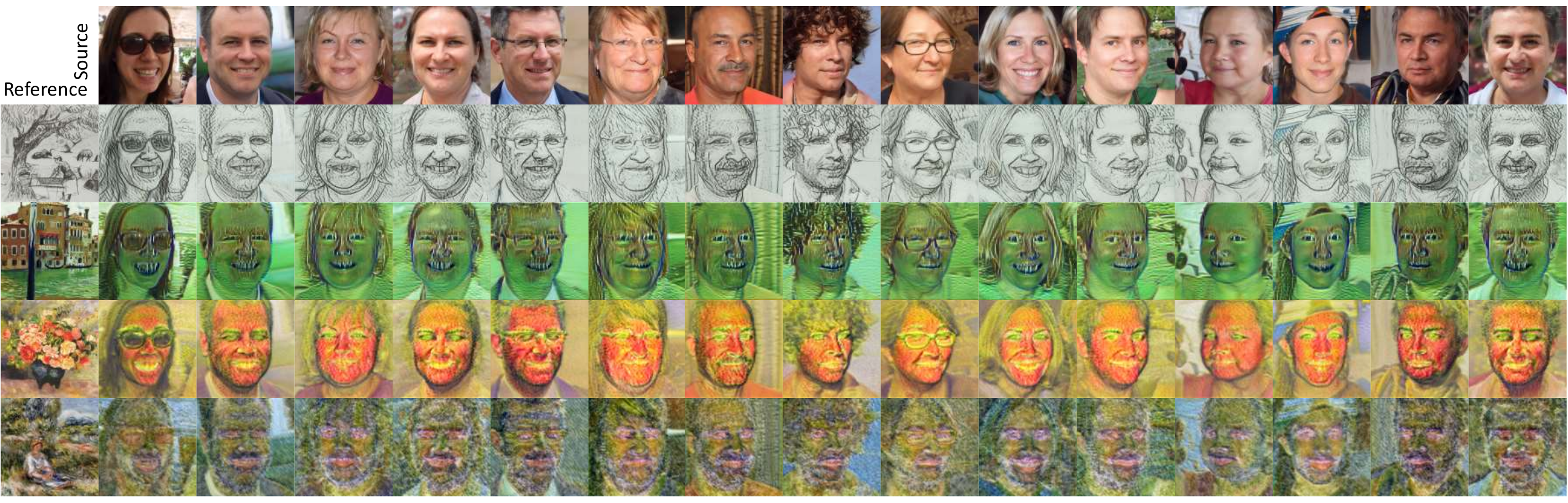}
	\caption{\textbf{Cross-domain adaptation} 3D-Adapter transfers the character of an out-of-domain target (first column) to the source domain. }
	\label{fig:eg3d_cross_domain}
\end{figure*}

In this section, we have conducted additional experiments to demonstrate cross-domain adaptation, similar to those presented in Figure 5 of One-Shot Generative Domain Adaptation \cite{yang2023one}. Specifically, we selected four natural images as reference images for a face source model, with the expectation that the synthesis after adaptation would maintain consistent visual concepts. In other words, a face model is anticipated to continue generating faces, regardless of the target image. As shown in Fig. \ref{fig:eg3d_cross_domain}, the source models successfully produce the corresponding content. However, due to the limited shared attributes between faces and natural images, the 3D-Adapter primarily focuses on variation factors such as color schemes, textures, and painting styles, which can be directly transferred across unrelated domains. Nonetheless, compared to domain adaptation within more closely related domains, cross-domain adaptation exhibits a decline in the quality of the generated images.



\section{More Comparison on One-shot 3D GDA}

The results in Figure 4 of the manuscript show that the adversarial-based method DoRM suffered from severe training failure, while the non-adversarial-based approach DiFa experienced model collapse, resulting in generated samples that fully replicate the training data and lack generative diversity.

\begin{figure*}[t!]
	\centering
	\includegraphics[scale=0.35]{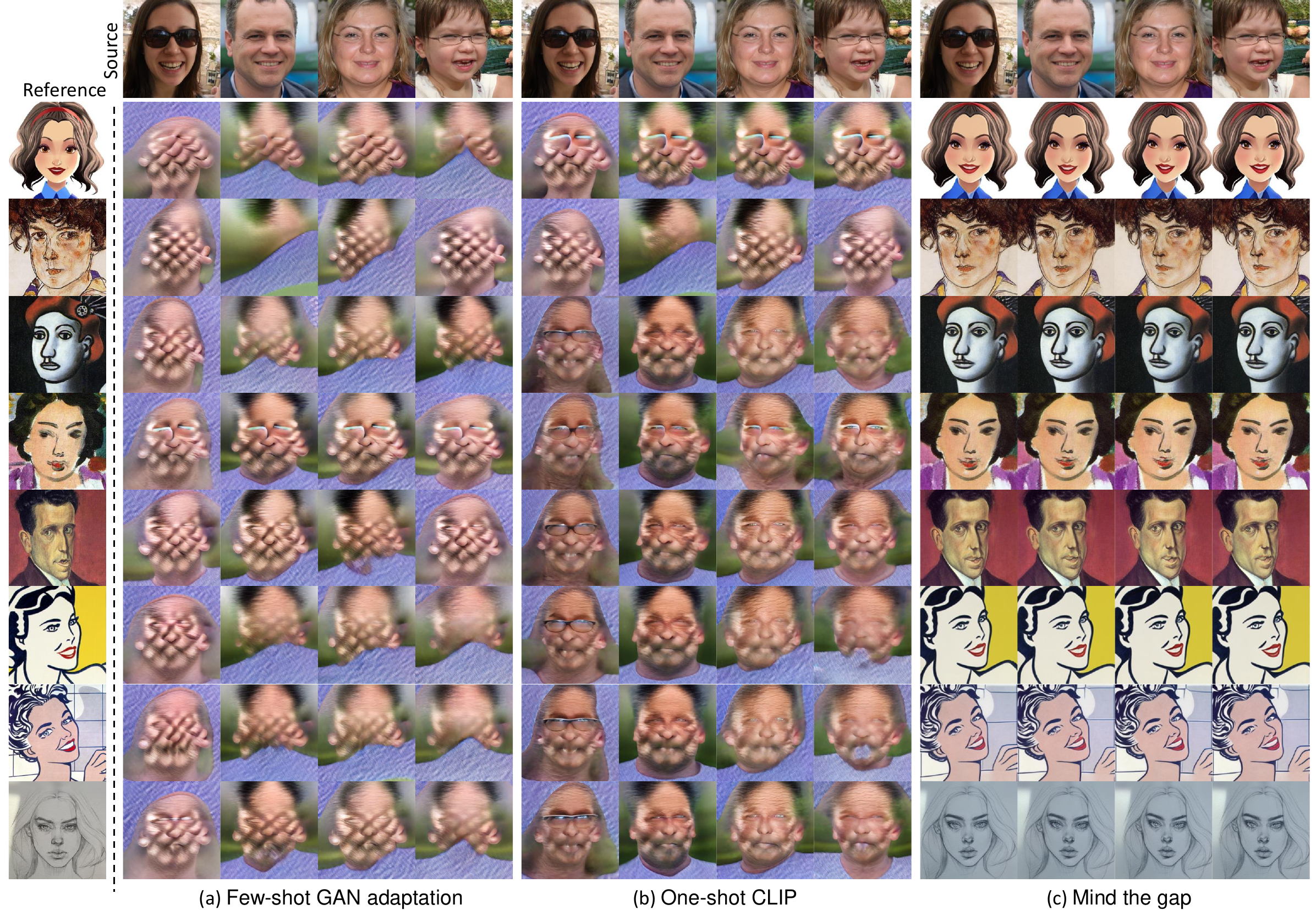}
	\caption{\textbf{Qualitative comparisons} on one-shot setting between One-shot CLIP (TPAMI 23) \cite{kwon2023one}, Few shot GAN adaptation (CVPR 21) \cite{ojha2021few}, and Mind the Gap (ICLR 22) \cite{zhu2021mind}. The first row and first column show different images in source domains and reference images in target domains. \textbf{Results best seen at 500\% zoom}. }
	\label{fig:one-shot_other}
\end{figure*}

To further verify our hypothesis, we have incorporated comparison results with additional baselines, including One-shot CLIP (TPAMI 23) \cite{kwon2023one}, Few shot GAN adaptation (CVPR 21) \cite{ojha2021few}, and Mind the Gap (ICLR 22) \cite{zhu2021mind}, in this section. The results, shown in Figure \ref{fig:one-shot_other}, confirm that adversarial-based methods like One-shot CLIP and Few-shot GAN Adaptation indeed suffer from severe training failures. Meanwhile, non-adversarial-based methods like Mind the Gap also experience model collapse, leading to generated samples that fully replicate the training data and lack generative diversity. By including these comparisons, we provide a more comprehensive evaluation of our proposed method’s performance and further substantiate its effectiveness.

\begin{figure*}[t!]

	\centering
	\includegraphics[scale=0.37]{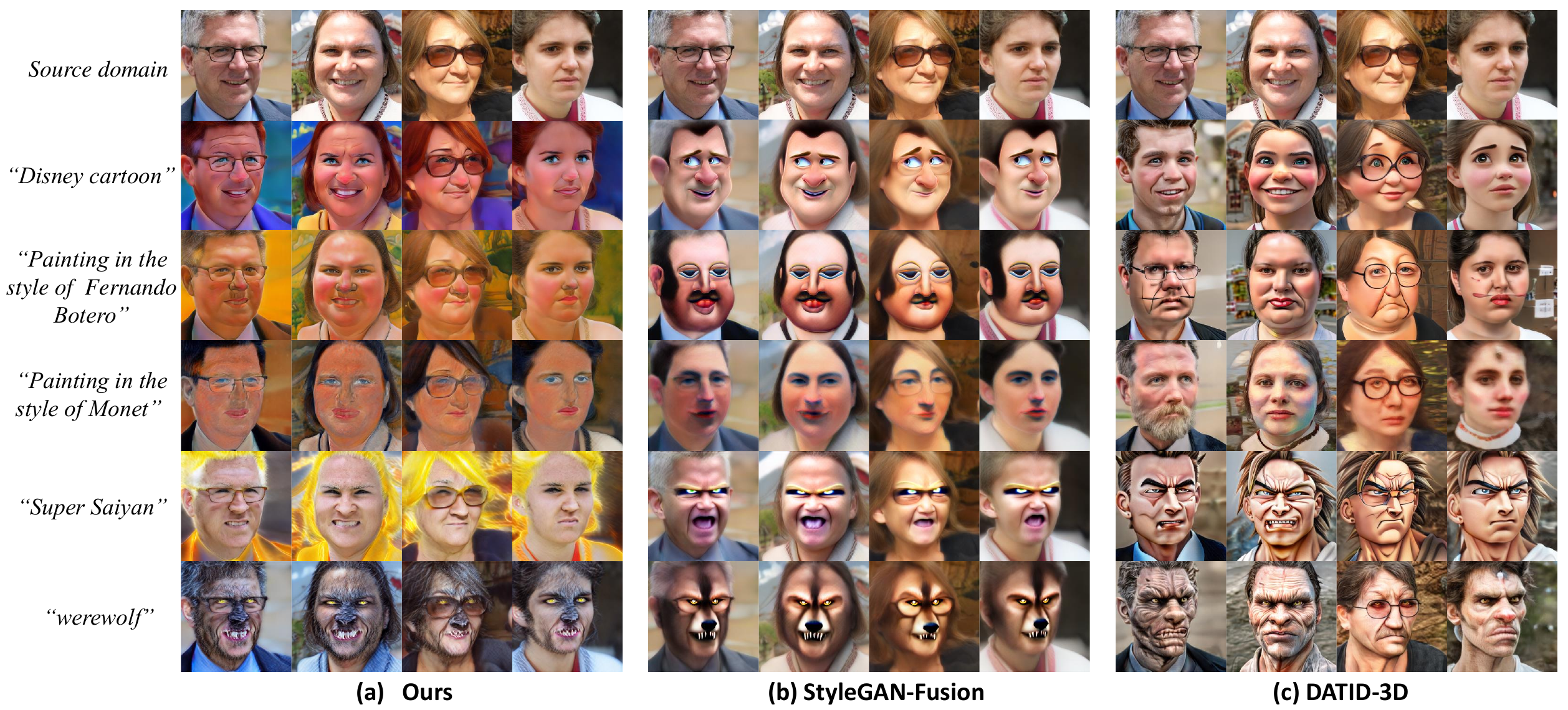}
	\caption{\textbf{Qualitative comparisons} on zero-shot setting between our proposed 3D-Adapter, StyleGAN-Fusion \cite{song2022diffusion} and DATID-3D \cite{kim2023datid}. The first row and first column show the source images and the descriptions of target domain, respectively. \textbf{Results best seen at 500\% zoom}. }
	\label{fig:zero-shot_R3}
\end{figure*}

\section{More Comparison on Zero-shot 3D GDA}

To provide a comprehensive evaluation, we have included a comparative analysis with StyleGAN-Fusion \cite{song2022diffusion} and DATID-3D \cite{kim2023datid}, as they are two relevant and accessible baselines. As shown in Figure \ref{fig:zero-shot_R3}, our proposed 3D-Adapter demonstrates superior fidelity, diversity, and cross-domain consistency compared to both StyleGAN-Fusion \cite{song2022diffusion} and DATID-3D \cite{kim2023datid}. Specifically, StyleGAN-Fusion \cite{song2022diffusion} and DATID-3D \cite{kim2023datid} exhibit lower fidelity and fail to retain certain domain-independent attributes of the source domain, such as gender and the presence of eyeglasses. Moreover, StyleGAN-Fusion \cite{song2022diffusion} and DATID-3D \cite{kim2023datid} encounter difficulties in adapting to the target domain under some settings.

\end{document}